\newif\ifarxiv
\arxivtrue

\documentclass[sn-mathphys-num]{sn-jnl}

\usepackage{graphicx}%
\usepackage{multirow}%
\usepackage{amsmath,amssymb,amsfonts}%
\usepackage{amsthm}%
\usepackage{mathrsfs}%
\usepackage[title]{appendix}%
\usepackage{xcolor}%
\usepackage{textcomp}%
\usepackage{manyfoot}%
\usepackage{booktabs}%
\usepackage{algorithm}%
\usepackage{algorithmic}%
\usepackage{listings}%

\usepackage{microtype}
\usepackage{graphicx}
\usepackage{subfigure}
\usepackage{booktabs} 
\usepackage{graphicx}%
\usepackage{multirow}%
\usepackage{amsmath,amssymb,amsfonts}%
\usepackage{amsthm}%
\usepackage{mathrsfs}%
\usepackage[title]{appendix}%
\usepackage{xcolor}%
\usepackage{textcomp}%
\usepackage{manyfoot}%
\usepackage{booktabs}%
\usepackage{listings}%
\usepackage{bbm}
\usepackage{soul}

\theoremstyle{thmstyleone}%
\theoremstyle{thmstyletwo}%

\theoremstyle{thmstylethree}%

\usepackage{letltxmacro}

\newlength{\commentindent}
\makeatletter
\renewcommand{\algorithmiccomment}[1]{\unskip\hfill\makebox[\commentindent][l]{//~#1}\par}
\LetLtxMacro{\oldalgorithmic}{\algorithmic}
\renewcommand{\algorithmic}[1][0]{%
  \oldalgorithmic[#1]%
  \renewcommand{\ALC@com}[1]{%
    \ifnum\pdfstrcmp{##1}{default}=0\else\algorithmiccomment{##1}\fi}%
}
\makeatother

\newcommand{\X}{\mathbf{X}}
\newcommand{\lv}{\mathbf{b}^l}
\newcommand{\uv}{\mathbf{b}^u}

\newcommand{\U}{\mathbf{U}}
\newcommand{\uu}{\mathbf{u}}
\newcommand{\n}{\mathbf{n}}
\newcommand{\Xc}{\mathcal{X}_c}
\newcommand{\Xcp}{\hat{\mathcal{X}_c}}

\newcommand{\Vc}{\mathcal{V}_{\mathcal{X}}}
\newcommand{\x}{\mathbf{x}}

\newcommand{\xt}{\tilde{\mathbf{x}}}
\newcommand{\z}{\mathbf{z}}
\newcommand{\y}{\mathbf{y}}
\newcommand{\xnew}{\tilde{\mathbf{x}}}

\newcommand{\e}{\mathbf{e}}

\begin{document}

\title[Voronoi Candidates]{Voronoi Candidates for Bayesian Optimization}


\author*[1]{\fnm{Nathan} \sur{Wycoff}}\email{nwycoff@umass.edu}

\author[2]{\fnm{John W.} \sur{Smith Jr.}}\email{john.smith20@montana.edu}

\author[3]{\fnm{Annie S.} \sur{Booth}}\email{anniees@vt.edu}

\author[3]{\fnm{Robert B.} \sur{Gramacy}}\email{rbg@vt.edu}

\affil*[1]{\orgdiv{Department of Mathematics and Statistics}, \orgname{University of Massachusetts}, \orgaddress{\street{Massachusetts Ave}, \city{Amherst}, \postcode{01003}, \state{MA}, \country{USA}}}

\affil[2]{\orgdiv{Department of Mathematical Sciences}, \orgname{Montana State University}, \orgaddress{P.O. Box 172400, Bozeman, \postcode{59717}, \state{MT}, \country{USA}}}

\affil[3]{\orgdiv{Department of Statistics}, \orgname{Virginia Tech}, \orgaddress{\street{250 Drillfield Dr.}, \city{Blacksburg}, \postcode{24061}, \state{VA}, \country{USA}}}


\abstract{
Bayesian optimization (BO) offers an elegant approach for efficiently optimizing black-box functions by sequentially choosing the most favorable point according to an acquisition criterion.
However, acquisition criteria demand their own challenging inner-optimization,
which can induce significant overhead. 
Many practical BO methods, particularly in high dimension, eschew a formal, continuous optimization of the acquisition function and instead search discretely over a finite set of candidates which are in some sense representative. 
Here we propose candidates which lie on the boundary of the Voronoi tessellation of the current design points, such that they are equidistant to two or more of them.
We discuss strategies for efficient implementation by directly sampling the boundary without explicitly generating the tessellation, thus accommodating large designs in high dimension.
On a battery of test problems optimized via Gaussian processes with expected improvement, our proposed approach demonstrates significantly reduced execution time relative to a multi-start continuous search while retaining or even improving accuracy on most examined test problems.
This has the potential to expand the class of problems on which BO is feasible.
}

\keywords{black-box, surrogate, emulator, derivative-free optimization, high dimension, computational geometry}



\maketitle




\section{Introduction}\label{sec:intro}

Bayesian Optimization \citep[BO; e.g.,][]{frazier2018tutorial} is the methodology of choice for global optimization of black-box functions of moderate dimension where each function evaluation comes at enormous cost.
This emphasis on expensive functions justifies the significant computational expenditure required to conduct BO, which in its classical formulation requires the analyst to solve an entire nonconvex optimization problem for each black-box function evaluation; we'll refer to this as the \textit{acquisition subproblem}.
It also means that performance of an optimizer under restrictive budgets of black-box evaluations, e.g., tracking the best observed value as budgets are increased, is often of greater
interest than other considerations, like eventual convergence.

To inform acquisitions, a statistical emulator or ``surrogate'' \citep[e.g.,][]{gramacy2020surrogates} model is trained on a limited set of observations from the black-box function.
A Gaussian process \citep[GP; e.g.,][]{williams2006gaussian} surrogate is 
often preferred for its predictive prowess and 
well-calibrated uncertainty quantification at unobserved inputs, though the GP itself rapidly grows expensive to evaluate as dataset size increases.
Together these feed into an acquisition function \citep{merrill2021empirical} which quantifies the utility of evaluating the black-box at a new input.
Subsequent evaluations are conducted by iterating through solving the acquisition subproblem, performing a black-box function evaluation, and updating the surrogate.

In low dimension, the acquisition subproblem can be globally solved by evaluating the acquisition function over a dense grid, but this quickly becomes infeasible as dimension grows.
Consequently, a common default for optimizing acquisition functions is a multi-start derivative-based numerical search.
It is crucial that the computational cost of solving the acquisition subproblem remain tractable so as to not overshadow the cost of function evaluations themselves.
This computational restriction relegates the elegant BO approach  to black-box problems with relatively small budgets and staggering function evaluation costs. 
But this excludes many interesting problems, including a class of statistical inference problems in ecology that motivated this work.

To address this issue, researchers have proposed diverse methods to improve the execution time of BO. 
One is to approximate the GP or other surrogate itself, so as to make the acquisition subproblem easier insofar as the underlying function and gradient evaluations are faster.
Examples include modeling a GP locally \cite{gramacy2015local},
imposing conditional independence \cite{vecchia1988estimation,katzfuss2020vecchia},
imposing sparsity in the covariance \cite{furrer2006covariance},
using finite-dimensional structure \cite{rahimi2007random},
and using inducing variables either alone \cite{williams2000using,snelson2005sparse} 
or as part of variational approximations \cite{titsias2009variational,hensman2013gaussian}. 
But the exponential growth of the search space as dimension increases limits the feasibility of these tactics for larger problems.
Another approach, the focus of this work, is to limit acquisitions to a discrete set of candidate locations \citep[e.g.,][]{kandasamy2018parallelised,eriksson2019scalable,gramacy2022triangulation}. 
This involves abandoning the idea of finding the global optimum of the acquisition surface 
using a continuous solver (which is an inherently sequential endeavor)
and instead converting the problem into a discrete and highly parallelizable calculation. 
If the global optimum of the acquisition subproblem happened to lie in the candidate set, this discretization would retain the same accuracy as a global continuous search.
In practice, this is too much to ask of a candidate set: we instead aim simply to have at least some candidates which obtain reasonably good acquisition values.

\begin{figure}[ht!]
    \centering
    \includegraphics[width=0.80\textwidth]{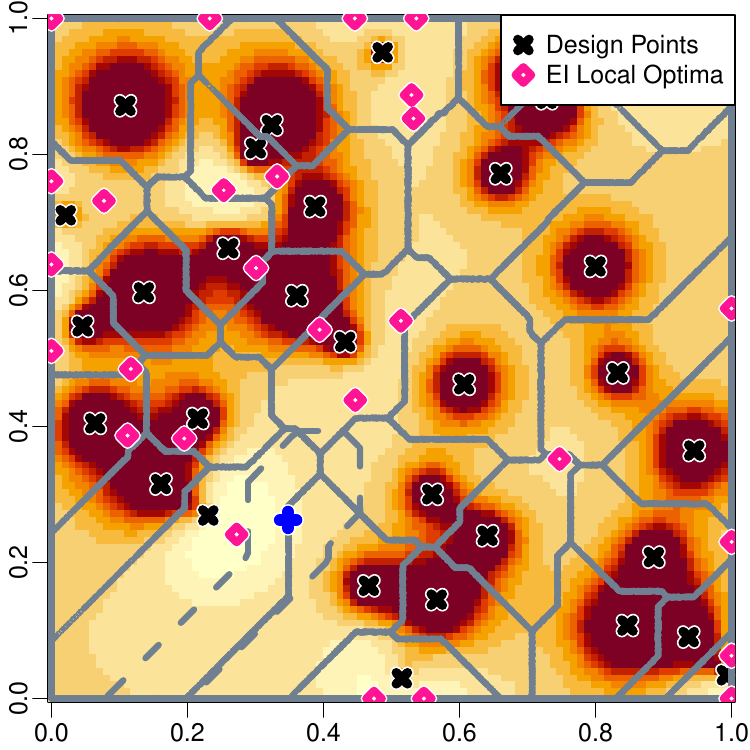}
    \includegraphics[width=0.145\textwidth]{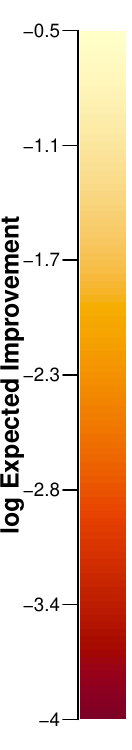}
    \caption{EI surface of a toy problem and its local optima with $\ell_\infty$ Voronoi tesselation (gray lines; see Section \ref{sec:bg_vor}) superimposed.
    Dotted gray line gives new boundary after adding the blue ``+".
    }
    \label{fig:expo}
\end{figure}

In this article, we will develop a design-dependent candidate scheme which is computationally tractable in high dimension. 
We operationalize the intuition that the optimum of the acquisition function is ``between'' existing design points \citep{gramacy2022triangulation} by proposing candidates which are equidistant to two or more of them. 
Figure \ref{fig:expo} illustrates our approach in two dimensions on the toy function $f(\x) = \sum_{p=1}^2 \sin\big(4\pi(x_p-0.5)^2\big)$ with the expected improvement \citep[EI;][]{jones1998efficient} acquisition criterion based on a GP surrogate fit.
Most of the local optima of EI (pink diamonds) happen to lie quite close to the set of equidistant points (gray lines, which we call the \textit{Voronoi boundary}), which itself tends to follow the ridges of high EI (yellow surface).
Optima which are not near the boundary can be reached in subsequent iterations as the Voronoi tesselation is refined (e.g., dotted gray line).

While explicit computation of an entire Voronoi tesselation is intractable in even moderate dimension, recent developments (see Section \ref{sec:bg_vorhigh}) allow us to efficiently identify individual points on the Voronoi boundary by working only implicitly with the tesselation.
Our method is bassed on nearest neighbor calculations, making implementation simple and computationally efficient.
There is considerable freedom in the method of discretizing the continuous Voronoi boundary.
We study several implementations, finding one of them to consistently offer equivalent or better best observed function values compared to a multi-start numerical EI search in a small fraction of the time. 
On our ecological test problems, we gain an order of magnitude improvement in execution time while 
even slightly improving on the best observed function values offered by the classical approaches.

Our paper is organized as follows.
In Section \ref{sec:bg} we discuss existing candidate approaches and show how they do not adequately scale to high dimensions.
In Section \ref{sec:meth} we introduce Voronoi candidates (or \textit{Vorcands}) for BO in high dimension and propose several concrete implementations.
In Section \ref{sec:exp}, we compare these implementations to classical approaches, finding huge gains in execution speed and more modest ones
in BO progress over restrictive budgets.
Finally, we comment on future work and offer substantive conclusions in Section \ref{sec:disc}.

\section{Existing Methods and their Limitations}\label{sec:bg} \label{sec:bg_bo}

We seek to minimize a deterministic function $f:[0, 1]^P \rightarrow \mathbb{R}$.  
The Bayesian approach to optimization \citep[for overviews see, e.g.,][]{garnett2023bayesian,shahriari2015taking,frazier2018tutorial} consists of placing some distribution on $f$ and developing a posterior conditioned on a set of observations. 
Specifically, we acquire a set of inputs $\X\in\mathbb{R}^{N\times P}$, with row $\x_i^\top$ representing a single design point, and observe the corresponding outputs $\y\in\mathbb{R}^N$, the smallest of which is $y_{\min}$.
The posterior distribution is treated as a \textit{surrogate model}: an approximation to $f$ which is cheap to evaluate, can interpolate observed input--output pairs, and provides reasonable uncertainty about unseen function values.
At each iteration, we use this posterior to decide on a promising next point $\xt$, evaluate $f(\xt)$, and then fold the new input--output pair $\{\xt,f(\xt)\}$ into $\X$ and $\y$. 
Typically, we begin with a small initial design before iterating until we exhaust a predetermined budget $B$. 
We denote the predictive mean and standard deviation of a surrogate model at location $\xnew$ as $\mu(\xnew)$ and $\sigma(\xnew)$.

The most commonly used surrogate model is a constant mean GP regression. 
For any $\X$, a GP induces a joint normal distribution $\mathbf{y}\sim \mathcal{N}(\mathbf{0},\mathbf{K})$, after centering, where $\mathbf{K}$ is a matrix with element $i,j$ given by a kernel function $k(\x_i,\x_j)$.
There are many options for $k$; in our numerical experiments we use the standard squared exponential kernel:
$$k(\x_i,\x_j) = \exp\left(- \sum_{p=1}^P (x_{i,p}-x_{j,p})^2/\ell_p\right),$$ 
where lengthscales $\boldsymbol\ell\in\mathbb{R}^P_+$ control the distance over which outcomes are correlated.  Conditional
on hyperparameters like $\boldsymbol\ell$, Gaussian predictive moments $\mu(\xnew)$ and $\sigma(\xnew)$ are available
in closed form.  We do not review these details here, as they are relatively well-established \citep[e.g., see][]{gramacy2020surrogates}
and there are many variations that work well and have been encapsulated in software libraries.
Moreover, although we exclusively employ GPs, none of our methodology depends on this choice, and the algorithms proposed are totally agnostic to the surrogate used. 
Rather, our metaheuristic can apply to any surrogate which has increasing uncertainty away from existing design points. 

One popular way of deciding on a new point $\xt$ is to optimize an \textit{acquisition function}, a scalar function $\alpha$ on $[0,1]^P$ which is determined by the posterior distribution on $f$, via $\mu(\xnew)$ and $\sigma(\xnew)$.
The most popular acquisition function is EI \citep{jones1998efficient,zhan2020expected} which synthesizes three crucial pieces of information: magnitude of improvement $\mu(\xnew) - y_{\min}$,
the probability of that improvement, and overall uncertainty about function outputs $\sigma(\xnew)$ at novel locations $\x$.  
Although we shall single out EI throughout our narrative for specificity, we see it as representative of a wide class of acquisition criteria $\alpha(\xnew)$ with similar properties and challenges, especially in high input dimension.  

In order to choose the next point to add to our design, we must solve the \textit{acquisition subproblem}, introduced in the previous section and formalized as follows:
\begin{equation}\label{eq:acq}
 \xt = \underset{\x\in[0,1]^P}{\textrm{argmax}}\, \; \alpha(\x)\,\,.
\end{equation}
Practical experience has found that this sub-problem exhibits pathological behavior such as regions of numerically zero gradient and large global variation in gradient norm.
\citet{ament2023unexpected} propose a method to alleviate the former issue of numerical precision for EI and some related acquisition functions, but the issue remains for more sophisticated acquistions.
The second issue, that the EI surface is nonconvex and exhibits many local optima (as illustrated in Figure \ref{fig:expo}), remains a significant challenge (though note that \citet{franey2011branch} have proposed a branch-and-bound algorithm).
Combined, this means that a serious multi-start search is necessary if one wants a real chance of finding the global optimum.


Even for a quick to evaluate surrogate, Problem \ref{eq:acq} can be quite challenging, which has lead some authors \citep[e.g.,][]{eriksson2019scalable,wang2020learning,eriksson2021scalable,daulton2022multi,gramacy2022triangulation} to abandon continuous optimization in favor of a finite candidate search. 
These methods proceed by defining some finite set $\Xc\subset [0,1]^P$, and selecting the next point as the discrete maximum over this set, as encapsulated in Algorithm \ref{alg:cands}.  This methodology can be applied to more sophisticated surrogates and acquisition criteria which do not have $\alpha(\x)$ or its derivatives available in closed form, as long as reasonable estimates are available.

\begin{algorithm}
    \caption{Bayesian Optimization with Candidates}
    \label{euclid}
    \begin{algorithmic}[1] 
            \FOR{$b \in \{1, \ldots, B\}$}
                \STATE $\X_c \gets C(\X, \y)$ \COMMENT{Propose candidates.}
                \STATE $\xt \gets \underset{\x\in\X_c}{\textrm{argmax}} \,\, \alpha(\x,\X,\y)$ \COMMENT{Evaluate discrete maximum.}
                \STATE $\X \gets [\X, \xt]$; $\y \gets [\y, f(\tilde{\x})]$
                \COMMENT{Augment design.}
            \ENDFOR
    \end{algorithmic}
    \label{alg:cands}
\end{algorithm}

\begin{figure}[ht!]
    \centering
    \includegraphics[width=0.48\textwidth]{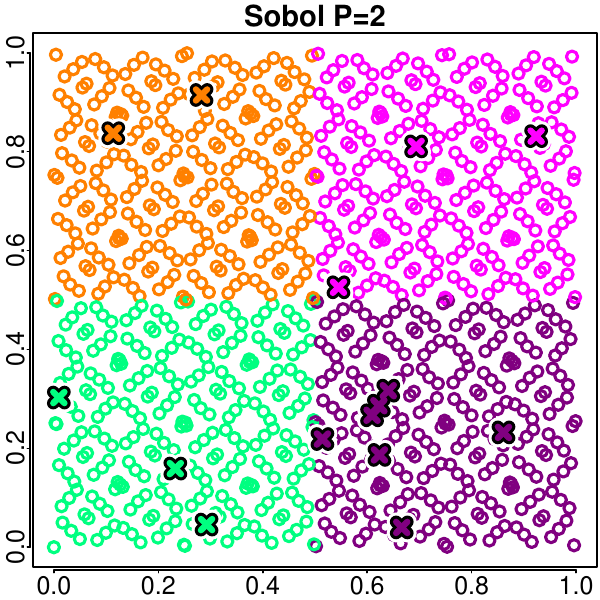}
    \includegraphics[width=0.48\textwidth]{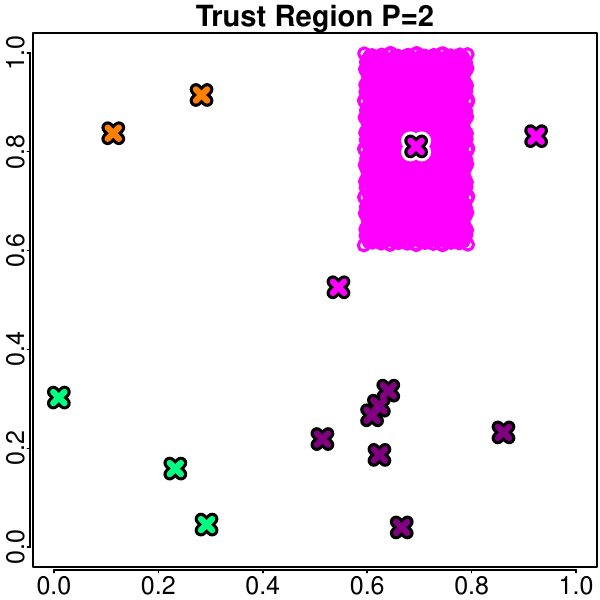}
    \includegraphics[width=0.48\textwidth]{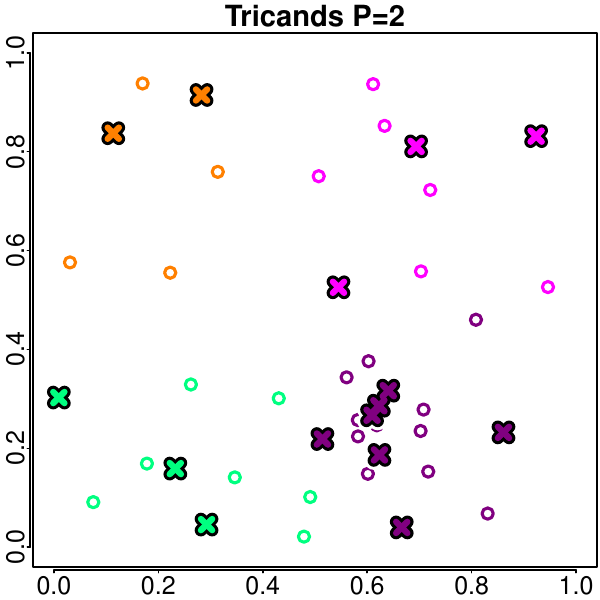}
    \includegraphics[width=0.48\textwidth]{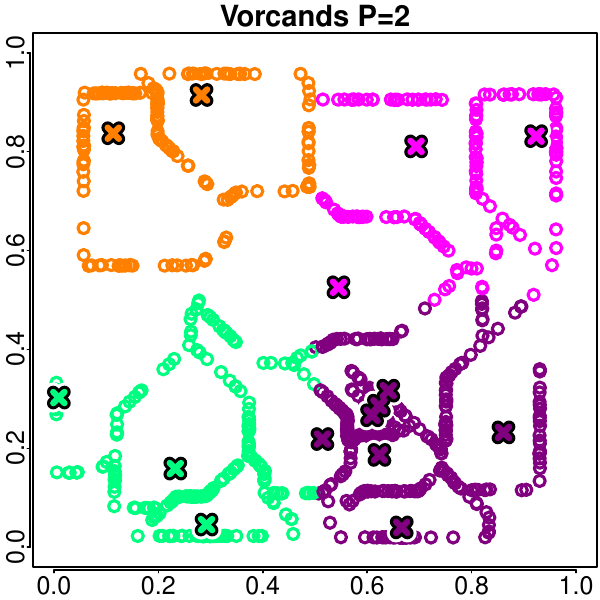}
    
    \caption{Comparing candidates for BO in 2 dimensions.  ``x''s represent $\X$ while open circles represent candidates. 
    Candidates in the same quadrant share a color.
    Tricands has a limited number of total candidates; all other figures show 1,000 candidates.}
    \label{fig:cands_compare_low}
\end{figure}

In low dimension, simple approaches such as setting $\Xc$ to be a Sobol sequence \cite{sobol1967distribution}, Latin hypercube sample \citep[LHS;][]{Mckay:1979} or other 
space-filling design \citep{lin2015latin} can be effective.  
Figure \ref{fig:cands_compare_low}, top left, illustrate that in low dimension, Sobol candidates (open circles) fill the space nicely.
There are about the same number of candidates in each quadrant, indicated by color.  
We cannot imagine that much would be lost if we restricted the acquisition subproblem from the entirety of $[0,1]^P$ to only look at the Sobol points.
Yet space filling candidates will struggle in higher dimensions -- more on that momentarily.

The challenge of higher dimension in part motivated ``Trust Region BO'' \cite[TuRBO;][]{eriksson2019scalable}, which restricts candidates to a GP lengthscale-stretched rectangle centered at the design point corresponding to the current optimum, as illustrated in Figure \ref{fig:cands_compare_low}, top right.
TuRBO is a sophisticated method which has much more to it than its candidate generation strategy, but that's what will be of interest to us in this article, and we'll refer to it as the Trust Region approach.
The candidates used are some space-filling design, originally a Sobol sequence, but restricted only to the trust region, which allows it to focus its candidates on a promising part of the space.
However, too small a trust region prevents exploration, while too large a trust region is no trust region at all.

More broadly, we'll refer to strategies such as trust regions, where $\Xc$ is computed on the basis of $\X$ and possibly $\y$, as \textit{data-dependent} candidate schemes. 
In contrast, the locations of the open circles $\mathcal{X}_c$ for the Sobol sequence (top left panel of Figure \ref{fig:cands_compare_low}) are chosen without consulting the colored ``x''s (representing the design points $\X$).
Another recent data-dependent scheme is triangulation candidates \citep{bates2001tri,gramacy2022triangulation}.
These ``Tricands'' build a Delaunay triangulation of the current design matrix $\X$, and choose centroids of this triangulation as elements of $\Xc$; see the lower left panel of Figure \ref{fig:cands_compare_low}. 
Each candidate point is the average of $P+1$ design points which are adjacent according to the triangulation. 
This allows Tricands to function effectively as an adaptive grid, becoming denser in parts of the space containing more design points.
For many surrogates, especially GPs, 
barycenters are near to where local predictive uncertainty $\sigma(\x)$ is maximized.
Unlike Sobol points or trust regions, there is a maximum number of possible Tricands, and this limit is easily reached in low dimension, as indicated by the relative paucity of candidates.

The lower right panel of Figure \ref{fig:cands_compare_low} previews our Vorcands approach. 
It is also data-adaptive like the Trust Region or Tricands methods. 
Though it covers every quadrant unlike the Trust Region, it does not try to fill all gaps in the space like the Sobol sequence.
And unlike Tricands, there is no limit to the possible number of Vorcands. 


\begin{figure}[ht!]
    \centering
    \includegraphics[width=0.49\textwidth]{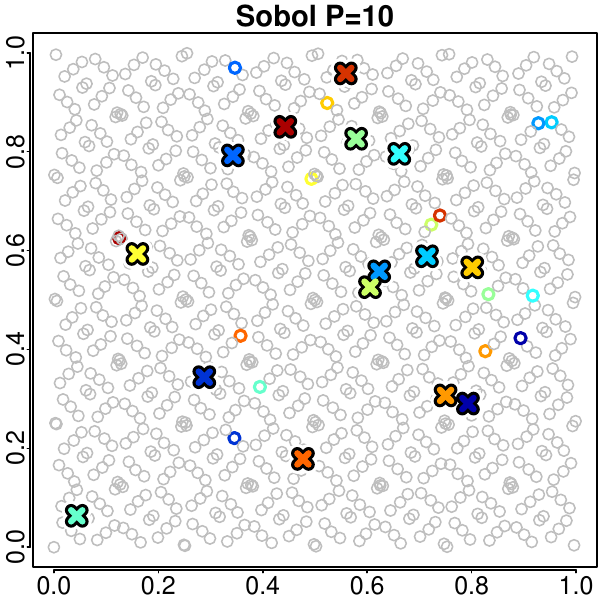}
    \includegraphics[width=0.49\textwidth]{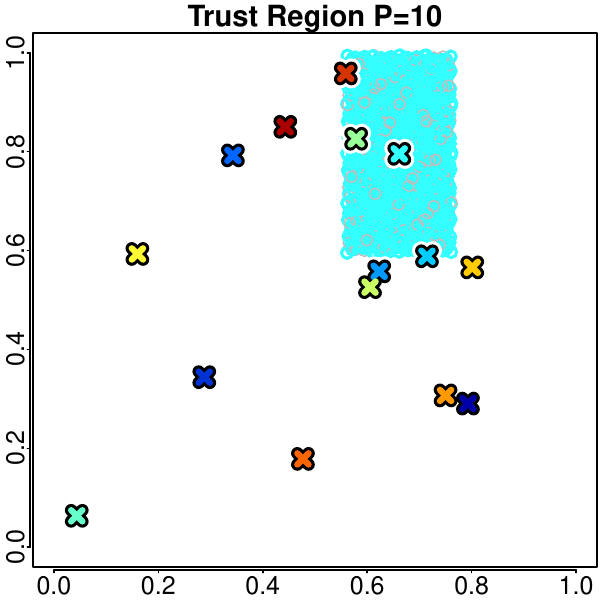}
    \includegraphics[width=0.49\textwidth]{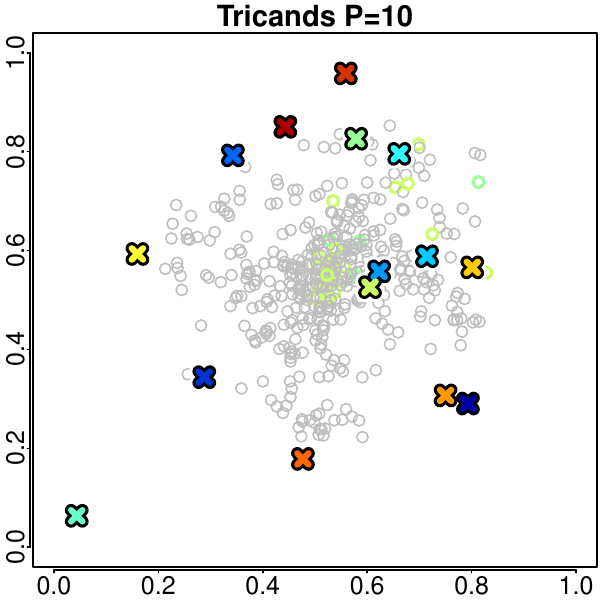}
    \includegraphics[width=0.49\textwidth]{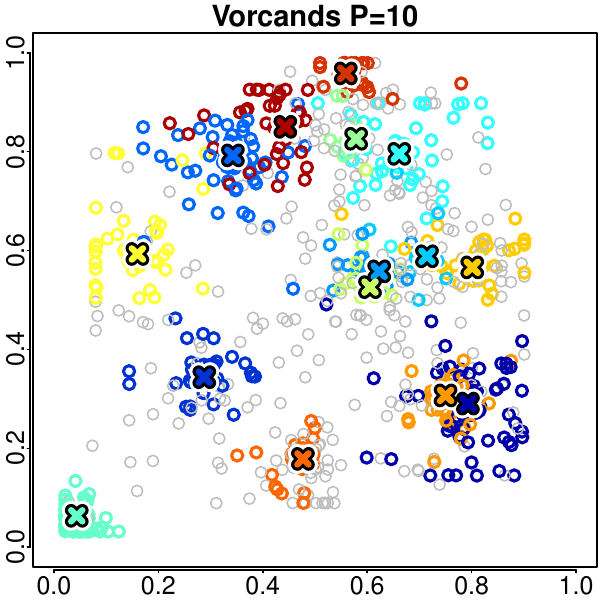}
    
    \caption{Comparing candidates for BO in 10 dimensions, the first two are shown.
    ``x''s represent $\X$ while open circles represent candidates. 
    Candidates in the same orthant share a color, except those not sharing an orthant with any design point, which are gray.
    All figures show 1,000 candidates.
}
    \label{fig:cands_compare_high}
\end{figure}

Now consider Figure \ref{fig:cands_compare_high}, showing a 2d projection of the same four candidate schemes in 10 dimensions.
Space-filling designs like Sobol sequences struggle to live up to their name in high dimension.
Although Sobol and LHS candidates have excellent projection properties, filling the entire volume requires exponential growth with input dimension $P$. 
For such candidates, only $\frac{1}{2^P}$ of its points will land anywhere in the orthant that the optimum lies in; even in ten dimensions, this means getting only one candidate on average in the appropriate orthant per $1{,}000$ candidate points.
The top-left panel of Figure \ref{fig:cands_compare_high} illustrates that, in 10 dimensions, the majority of Sobol candidates don't share an orthant (those shaded in gray) with \textit{any} design point (given by the ``x''s); LHS candidates behave similarly but are not pictured.  

The Trust Region approach behaves much as it did in low dimension when viewed via this projection, indicated by the striking similarity between the top right panels of Figures \ref{fig:cands_compare_low} and \ref{fig:cands_compare_high}.
The Trust Region proposes points in the same orthant as the current optimum but nowhere else.
Tricands, on the other hand, looks quite different in 10d. 
Things break down in two primary ways.
One is that Tricands collapse toward the middle, as seen in the lower left panel of Figure \ref{fig:cands_compare_high}.
Another is computational cost.
In higher dimension there are more simplices and exponentially more barycenters, meaning even the fastest algorithms \citep[e.g., quickhull,][]{quickhull} are painfully slow.
For instance, with $N=100$ and $P=10$, Tricands takes over two minutes to compute 2,000 candidates. 
This makes it hard to manage the effort involved in finding Tricands, and in controlling the expense of evaluating $\alpha(\cdot)$ on a potentially huge $\mathcal{X}_c$.
Although the scheme guarantees candidates $\mathcal{X}_c$ ``between'' the $\X$s, finding good ones in high dimension can be like looking for a needle in a (computationally expensive) haystack.

The next section describes our new data-dependent Vorcands candidate scheme, which, unlike Tricands, can be efficiently computed in high dimension (taking only seconds for a design of size 2,000 in dimension $P=100$) and avoids the mean-concentration phenomenon.
As shown in the bottom right panel of \ref{fig:cands_compare_high}, Vorcands provides points within the same orthant as current design points while also spreading reasonably throughout the 2d projection of the space, obtaining the desirable properties of both trust regions and space-filling designs.

\section{Voronoi Candidates}\label{sec:meth}

Our contribution may be summed up as simply proposing candidates on the boundary of a Voronoi tesselation of the training data, though implementing this in practice requires careful development.
First, there are multiple ways to define a Voronoi tesselation.
Second, and most importantly, identifying optimal discrete points on the Voronoi boundary is tricky; we propose two options and discuss their relevant merits.
We begin with a review of Voronoi tesselations to motivate these issues.

\subsection{The Voronoi tesselation}\label{sec:bg_vor}

Let $\X\subset [0,1]^P$ be a set of points and $d:[0,1]^P \times [0,1]^P\to\mathbb{R}^+$ be a dissimilarity function which is symmetric (i.e., $d(\x_1,\x_2) = d(\x_2,\x_1)$) and sub-additive (i.e., obeys the triangle inequality).
The Voronoi cell \citep[e.g.,][Chapter 10]{schneider2008stochastic} parameterized by $d$ associated with the $i^\mathrm{th}$ point is defined as the set of points whose dissimilarity with $\x_i$ is at least as small as their dissimilarity with any other point.  That is, 
$$\Vc^i = \{\x\in[0,1]^P : d(\x,\x_i) \leq d(\x, \x_j) \; \forall j \in \{1,\dots,N\}\}.$$
The Voronoi tesselation is the set of Voronoi cells belonging to all $i$.
The boundary of a given Voronoi cell, denoted by $\partial \Vc^i$, is of particular interest in this article: it consists of points which are equidistant to two or more points in $\X$ under $d$ (or are on the boundary of the unit hypercube).
We let $\partial\Vc := \cup_{i\in\{1,\ldots,N\}} \partial\Vc^i$, that is, $\partial\Vc$ denotes the set of all points which lie on at least one Voronoi cell's boundary.

\begin{figure}
    \centering
    \includegraphics[width=0.32\textwidth,trim=22 22 10 0,clip=TRUE]{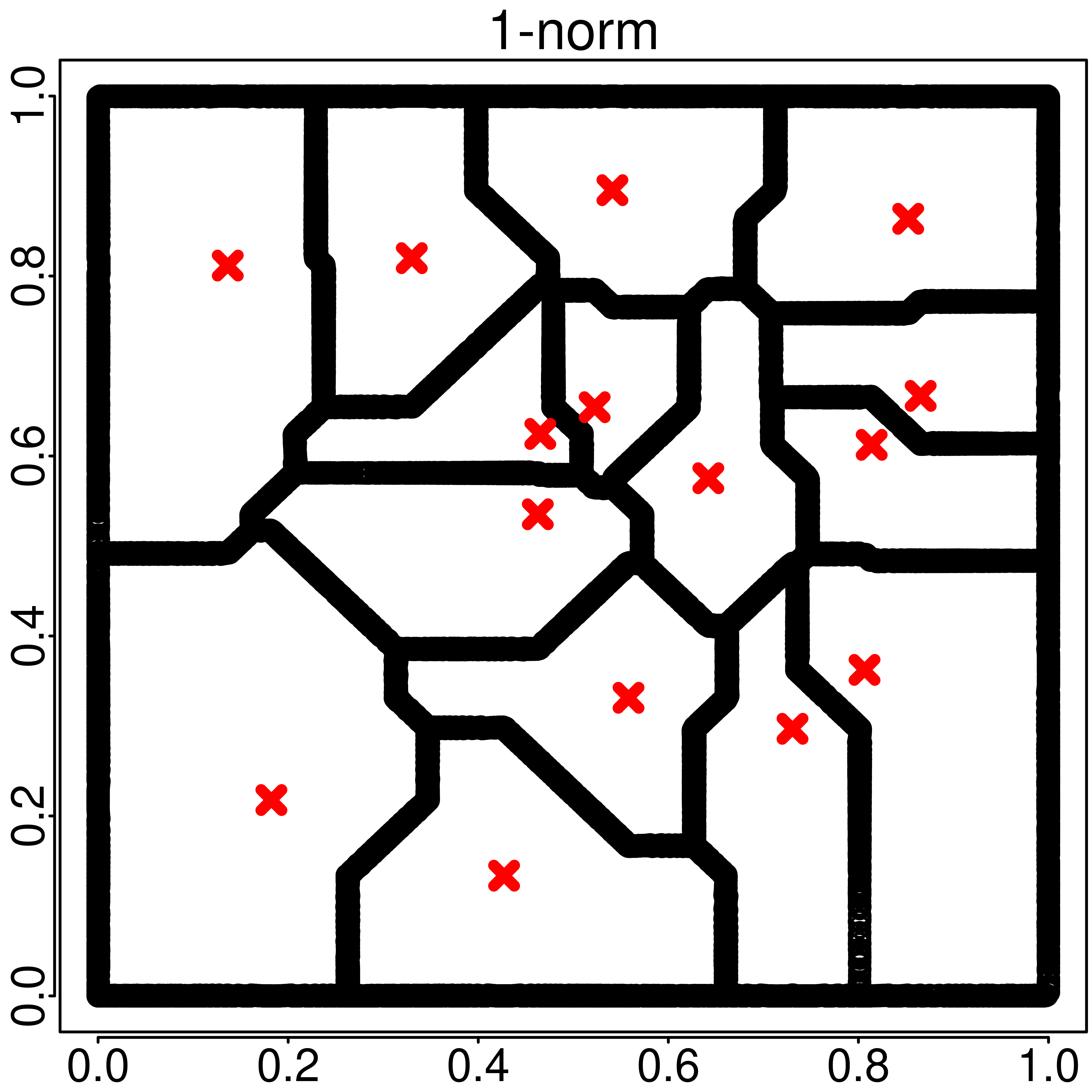}
    \includegraphics[width=0.32\textwidth,trim=22 22 10 0,clip=TRUE]{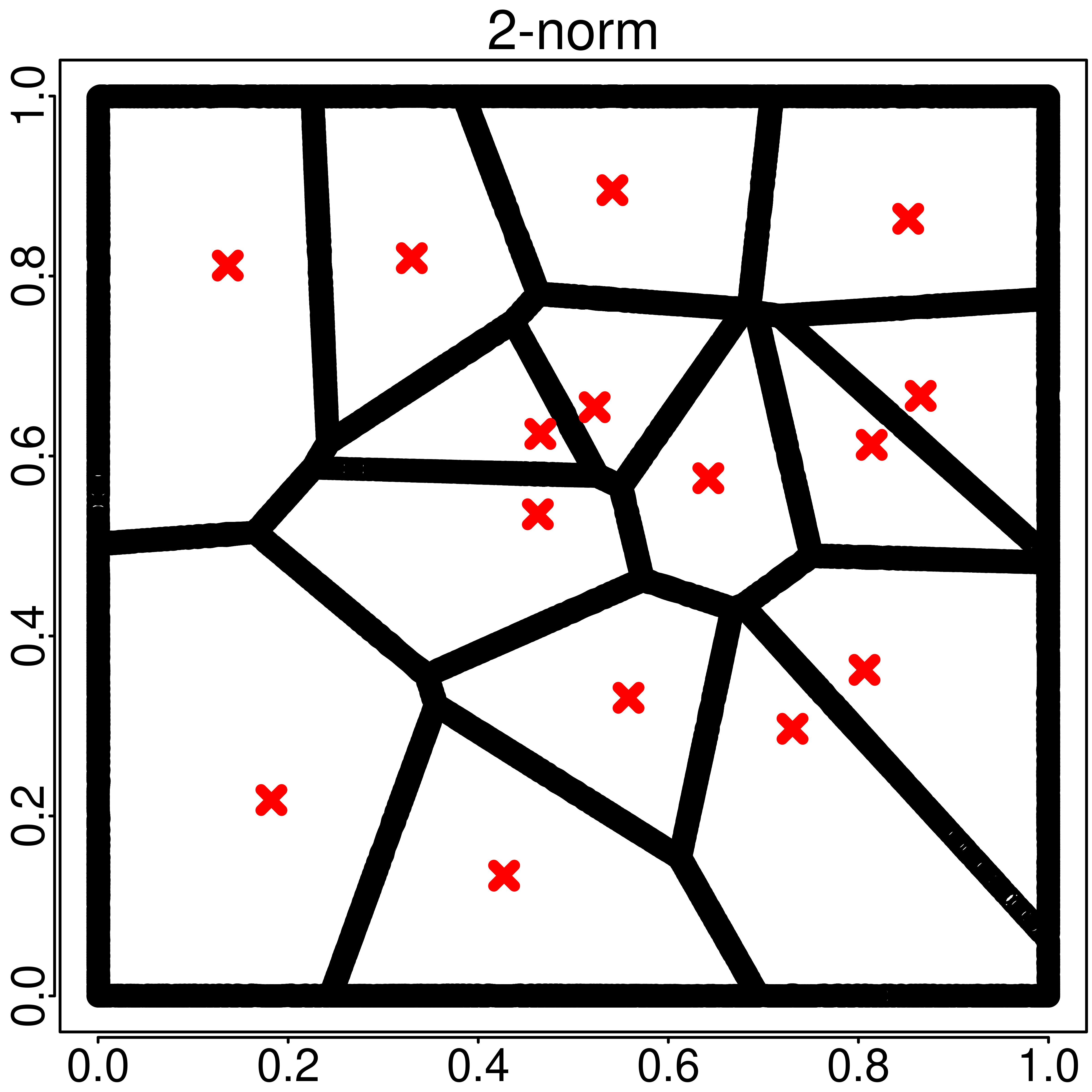}
    \includegraphics[width=0.32\textwidth,trim=22 22 10 0,clip=TRUE]{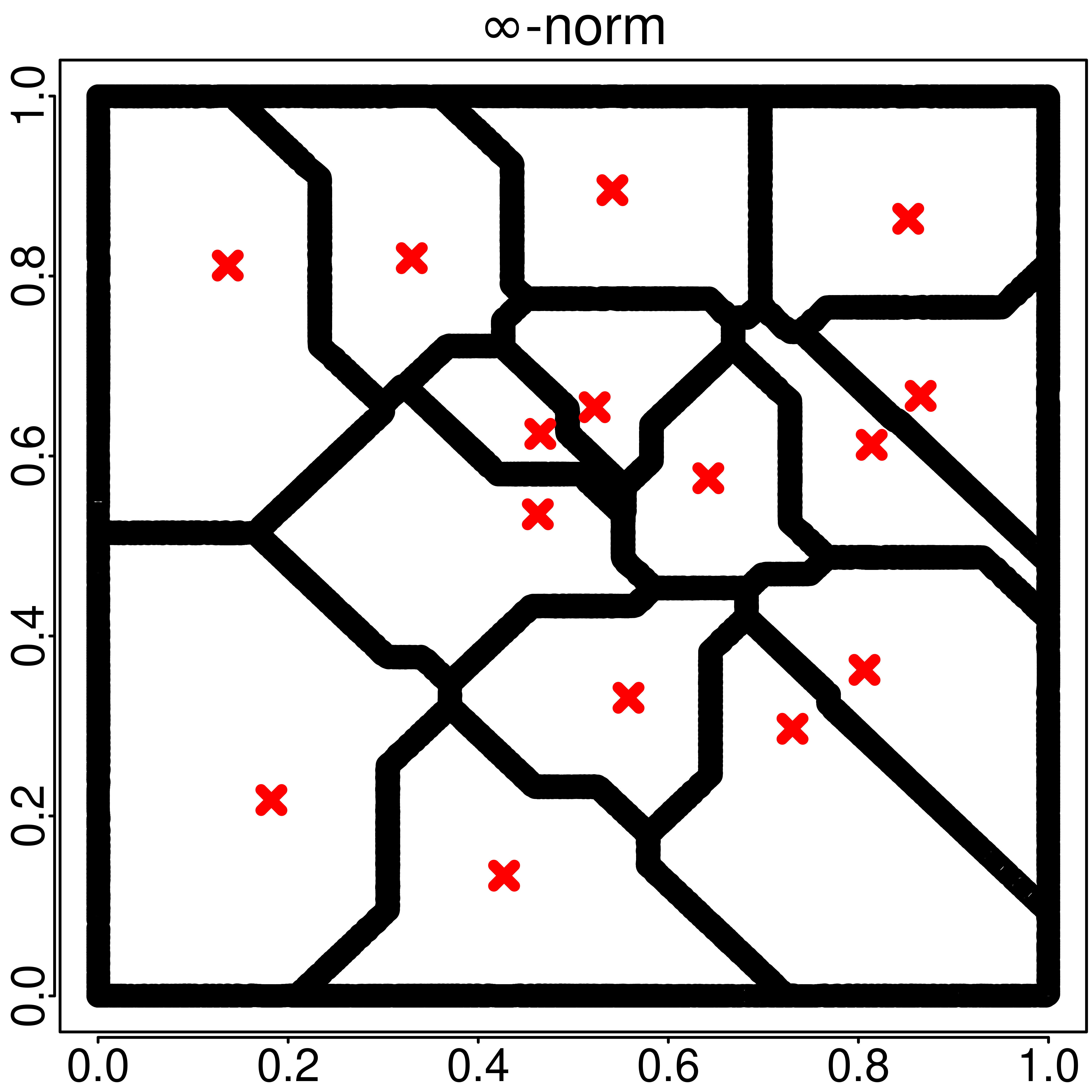}
    \caption{Example Voronoi cells induced by the 1, 2 and $\infty$ norm.}
    \label{fig:vornoi_illus}
\end{figure}

We restrict our attention to sub-additive dissimilarities as this is sufficient to ensure that the $i^{\mathrm{th}}$ Voronoi cell is star-convex with respect to $\x_i$, which is to say that for any point $\hat{\x}\in\Vc^i$, all of the points on the line segment connecting $\x_i$ and $\hat{\x}$ belong to $\Vc^i$ as well.
We will in particular focus on the dissimilarities given by the 1, 2 and $\infty$ norms where $d(\x_i,\x_j)$ is defined by
$$
\sum_{p=1}^P |x_{i,p} - x_{j,p}|,
\quad
\sqrt{\sum_{p=1}^P (x_{i,p} - x_{j,p})^2},
\quad \mbox{and} \quad
\max_{p} |x_{i,p}-x_{j,p}|
$$
respectively.
Figure \ref{fig:vornoi_illus} provides an illustration of Voronoi tesselations under those three dissimilarities for the same design $\X$, indicated by red ``x''s.

\subsection{Computation in High Dimension}
\label{sec:bg_vorhigh}

\begin{figure}[ht!]
    \centering
    \includegraphics[width=0.48\linewidth]{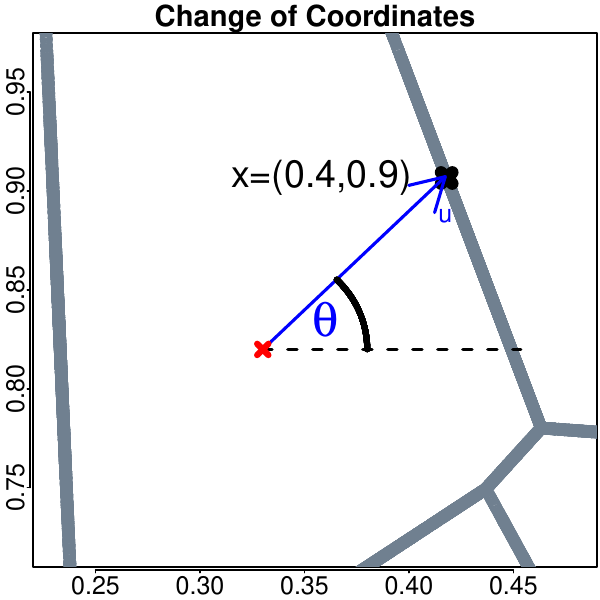}
    \includegraphics[width=0.48\textwidth]{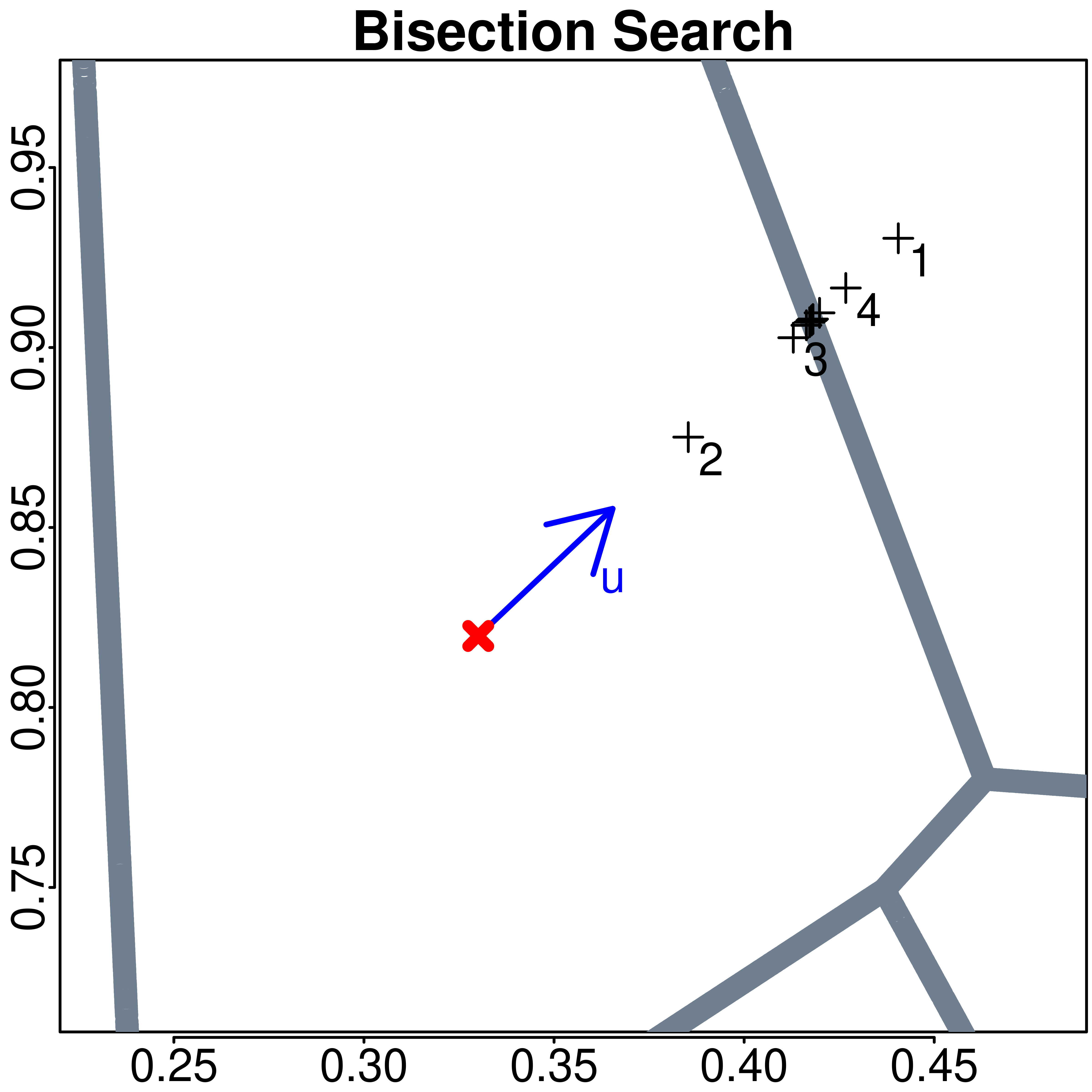}
    \caption{ As shown in the left panel, any point on the boundary of a Voronoi cell (black x) can be represented with Cartesian coordinates per usual, but may also be parameterized by the angle one has to travel (in blue) to reach it if starting at the design point defining the cell (red x).
    The right panel illustrates a bisection search to resolve the Cartesian coordinates of the implied boundary point given the starting point and the search direction.
    }
    \label{fig:coords}
\end{figure}

The problem of enumerating all the faces of a Voronoi tesselation is well studied and implemented in open source software, notably \texttt{quickhull} \cite{quickhull}.
However, generically, the number of faces of a Voronoi cell grows very quickly in $P$, rendering this approach infeasible in even moderate dimension. 
But to obtain a sample of points lying in $\partial\Vc$, we do not need to explicitly form the tesselation.
\citet{polianskii2022breaking} showed how we may rather work implicitly with Voronoi cells in high dimension by parameterizing any given point on the Voronoi boundary in terms of the center of the cell to which it belongs together with the direction we need to move from that cell center in order to reach the given point (see Figure \ref{fig:coords}). 
Consequently, in order to sample a point on the Voronoi boundary, we need only choose a starting design point $\x_i$ and a direction $\mathbf{u}$.
Then, we can find the distance $t$ that we must move, along the direction $\mathbf{u}$, to reach the boundary by conducting a bisection search to find the zero of the function 
\begin{equation}
 h(t) = d(\x_i,\x_i+t\mathbf{u}) - \underset{j\neq i}{\min} \,\, d(\x_j,\x_i+t\mathbf{u})\,\,.
\end{equation}
This can be solved implicitly using off-the-shelf nearest-neighbor software \citep[e.g.][]{arya1998optimal}: if the nearest neighbor to $\x_i+t\mathbf{u}$ in the design set is $\x_i$, then $t$ should be increased.
If it is some other point $\x_j$, it should be decreased.    
This procedure is illustrated in Figure \ref{fig:coords}, right.

\subsection{Acquisitions on the Voronoi Boundary}

We propose using points on the boundaries of Voronoi cells, that is, $\Xc \subset \partial\Vc$, as our candidate set for BO.
Since $\partial\Vc$ consists of uncountably many points, a practical method requires a strategy for producing some finite subset of it.
In this article, we study methods which achieve this by generating a random sample of points in $\partial\Vc$.
By the discussion in the previous section, we know that a sampling strategy on $\partial\Vc$ can be parameterized as a method for sampling 1) a Voronoi cell $n$ and 2) a direction $\uu$.

A given $n$ and $\uu$ imply some point on $\partial\Vc^n$; the Cartesian coordinates of that point may be determined using a bisection search as mentioned in the previous subsection.
We note that computing the coordinates of many such points at a time in order to form an entire candidate set is embarrassingly parallel.
However, modern nearest neighbor algorithms, such as that provided by {\tt ANN},\footnote{\url{https://github.com/dials/annlib}} can take advantage of batched calls for nearest neighbors (using a \textit{K-D tree} \cite{bentley1975multidimensional}) such that running a single loop and evaluating the neighborhood structure of all candidates simultaneously is more efficient. 
This approach is implemented in the following algorithm, where $\mathbf{n}\in\mathbb{N}^{C}$ gives the vector of cell indices and $\mathbf{U}\in\mathbb{R}^{C\times P}$ is a matrix containing a direction vector in each row.
The other parts of the procedure, such as updating bisection parameters and sampling search directions, can be computed in parallel, but we remark that the overall cost of the procedure is dominated by the nearest neighbors search.
We call this the ``Voronoi Walk'' as it effectively walks from each selected data point to the Voronoi boundary.


\begin{algorithm}[h]
    \caption{The Voronoi Walk (Vorwalk)}
    \label{alg:walk}
    \begin{algorithmic}[1] 
            \STATE {\bfseries Input:} $\X\in\mathbb{R}^{N\times P}, \n \in \{1, \ldots, N\}^C, \U\in\mathbb{R}^{C\times P}$ and $K$ (number of bisection iterations).
            \STATE $\lv \gets \mathbf{0}$; $\uv \gets \mathbf{1}$  \COMMENT{Initialize lower and upper bounds}
            \FOR{$k \in \{1, \ldots, K\}$} 
                \STATE $\mathbf{m} \gets \frac{\uv+\lv}{2}$ \COMMENT{Compute search interval midpoint}
                \STATE $\X_c \gets \X + \mathbf{m} \U$ \COMMENT{Scale search directions by $\mathbf{m}$.}
                \STATE $\mathbf{g} \gets$ near\_neigh$(\X, \X_c)$\COMMENT{Find Voronoi cell of candidates.}
                \FOR{$c \in \{1, \ldots, C\}$}  
                    \IF {$g_c==n_c$} 
                    \STATE $b^l_c \gets m_c $
                    \COMMENT{If inside origin cell, shrink $m_c$.}
                    \ELSE
                    \STATE $b^u_c \gets m_c $
                    \COMMENT{Else grow $m_c$.}
                    \ENDIF
                \ENDFOR
            \ENDFOR
            \STATE {\bfseries Return:} $\X_c$
    \end{algorithmic}
\end{algorithm}

Note that $\U$ must be scaled such that a step size of one will be guaranteed to lie outside of the pertinent Voronoi cell; this is guaranteed if $\Vert \mathbf{u}_i \Vert_2 > \sqrt{P}$ since it will actually lie outside of $[0,1]^P$.  We next discuss two meta-strategies for defining the distributions on $n$ and $\uu$.
\begin{figure}[ht!]
    \centering
    \includegraphics[width=0.5\textwidth]{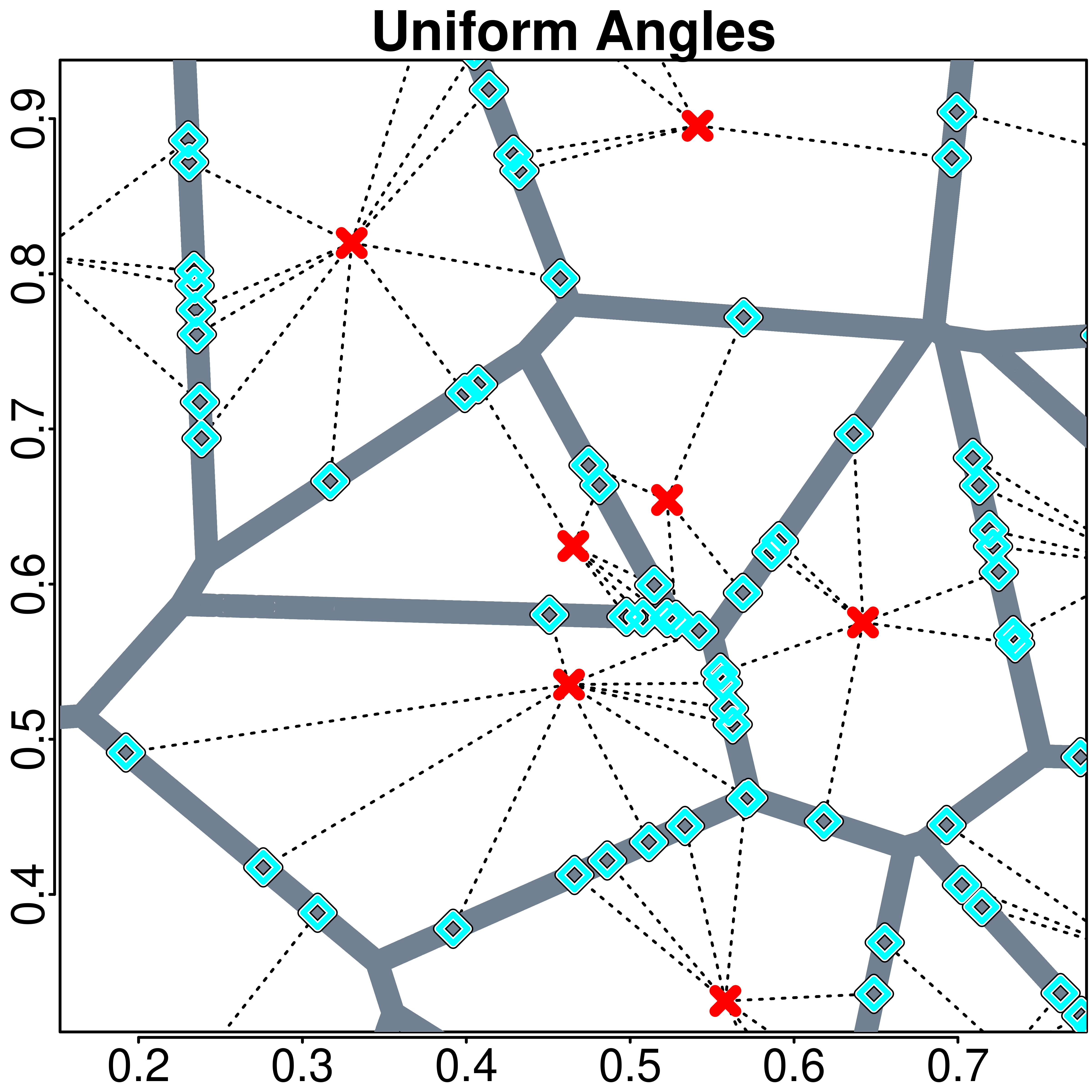}
    
    \caption{Voronoi candidates obtained from the \texttt{unif} strategy.
    Red ``x''s are design points; cyan diamonds are candidates. The gray lines give the Voronoi boundary.}
    \label{fig:walk}
\end{figure}

\subsubsection{Direct Sampling of $n$ and $\uu$}\label{sec:meth_ray}

This section studies the meta-strategy of directly sampling from a simple probability distribution on $n$ and $\uu$, denoted $\gamma$ and $\nu$, respectively. 
For instance, if we wish to concentrate on the part of the input space near existing design points, a natural choice would be to sample $n$ from a uniform distribution on $\{1,\ldots,N\}$.
This would ensure that our candidates consist of points adjacent in the tesselation of our design points $\X$, about equally.
Such a procedure may readily be modified to favor the Voronoi cells of certain points over others, say by reference to $\y$. 

Subsequently, $\uu$ is sampled from some simple distribution over the $P$-sphere.
We may modify any distribution over $\mathbb{R}^P$ to be a distribution over the $P$-sphere by dividing the sampled vector by its magnitude.\footnote{In order for this to be well defined that measure on $\mathbb{R}^P$ should assign probability zero to the origin.}
Treating an isotropic normal distribution in this manner yields a uniform distribution over the sphere.
We will refer to the strategy which uses a uniform distribution over both $n$ and $\uu$ as the \texttt{unif} sampling strategy.
We should point out that the uniform distribution on angles does not lead to a uniform distribution over the actual Voronoi boundary; points on boundaries closer to the design point defining a Voronoi cell will be overrepresented.

Conceptually, we could deploy standard simulation techniques, such as Markov-chain Monte Carlo, to produce a uniform sample of the boundary.
But the most straightforward implementation of this would lead to significantly more computational overhead, and we're in search of a thrifty method.

As an alternative to uniform sampling, we could consider sampling only $\uu$ which align with a coordinate axis, i.e., sampling uniformly from $\{\e_1,-\e_1,\ldots,\e_P,-\e_P\}$, where $\mathbf{e}_i$ denotes the $i^\textrm{th}$ coordinate vector.
Taken together with again a uniform measure on $n$, we call this the \texttt{rect} strategy.


Whether we are using the \texttt{unif} or \texttt{rect} sampling strategy, or any other $\gamma$ and $\nu$, we can develop $C$ many candidates using Algorithm \ref{alg:vorwalk} on the sampled cells and directions, as illustrated in Figure \ref{fig:walk}.
Of course, there would be no added difficulty to sampling dependent $n$ and $\uu$; this remains as interesting future work.

\begin{algorithm}[h]
    \caption{Direct Sampling of Candidates}
    \label{alg:vorwalk}
    \begin{algorithmic}[1] 
            \STATE {\bfseries Input:} $\X, \nu, \gamma, C$ \COMMENT{C is the number of candidates}
            \FOR{$c \in \{1, \ldots, C\}$}  
                \STATE $n[c] \sim \gamma$ \COMMENT{Sample Voronoi cell}
                \STATE $\U[c,:] \sim \nu$ \COMMENT{Sample direction to walk}
            \ENDFOR
            \STATE {\bfseries Return:} Vorwalk($\X, \n, \mathbf{U}$)
            \COMMENT{Algorithm 2}
    \end{algorithmic}
\end{algorithm}

Taking stock, we have proposed a means of subsampling $[0,1]^P$ which is computationally efficient and can adapt to the density of the data.
In the context of BO, this adaptation allows the sampling to both exploit places near the current best point, which likely has several nearby evaluated design points, and explore the remaining space.


\subsubsection{Implicit Sampling by Projection}\label{sec:meth_proj}

The meta-strategy of the previous showed how to use the Voronoi tesselation to generate candidates \textit{de novo}.
But there already exist many schemes for space filling candidates, and a natural question is whether they can be incorporated into the Vorcands framework.
This is the question we address in this section.
We do so by considering an implicitly defined a joint distribution on $n$ and $\uu$ given by sampling a point $\mathbf{z}\in[0,1]^P$, which we'll call a \textit{precandidate}.
We wish to associate with $\z$ some point on the Voronoi boundary.
A natural idea would be to compute the Euclidean projection of $\z$ onto $\partial\Vc$. 
However, the Voronoi boundary is highly nonconvex, as the interior of the Voronoi cells form massive gaps. 
Consequently, Euclidean projection of a point onto the boundary consists of a non-convexly constrained quadratic program.
In small dimension, all faces of the boundary of a given Voronoi cell could be enumerated.  Since each is individually a convex object, we could solve the projection onto each one and then take the face with the smallest residual.
But this quickly becomes computationally infeasible in high dimension due to the explosion in the number of faces.

\begin{figure}[ht!]
    \centering
    \includegraphics[width=0.48\textwidth]{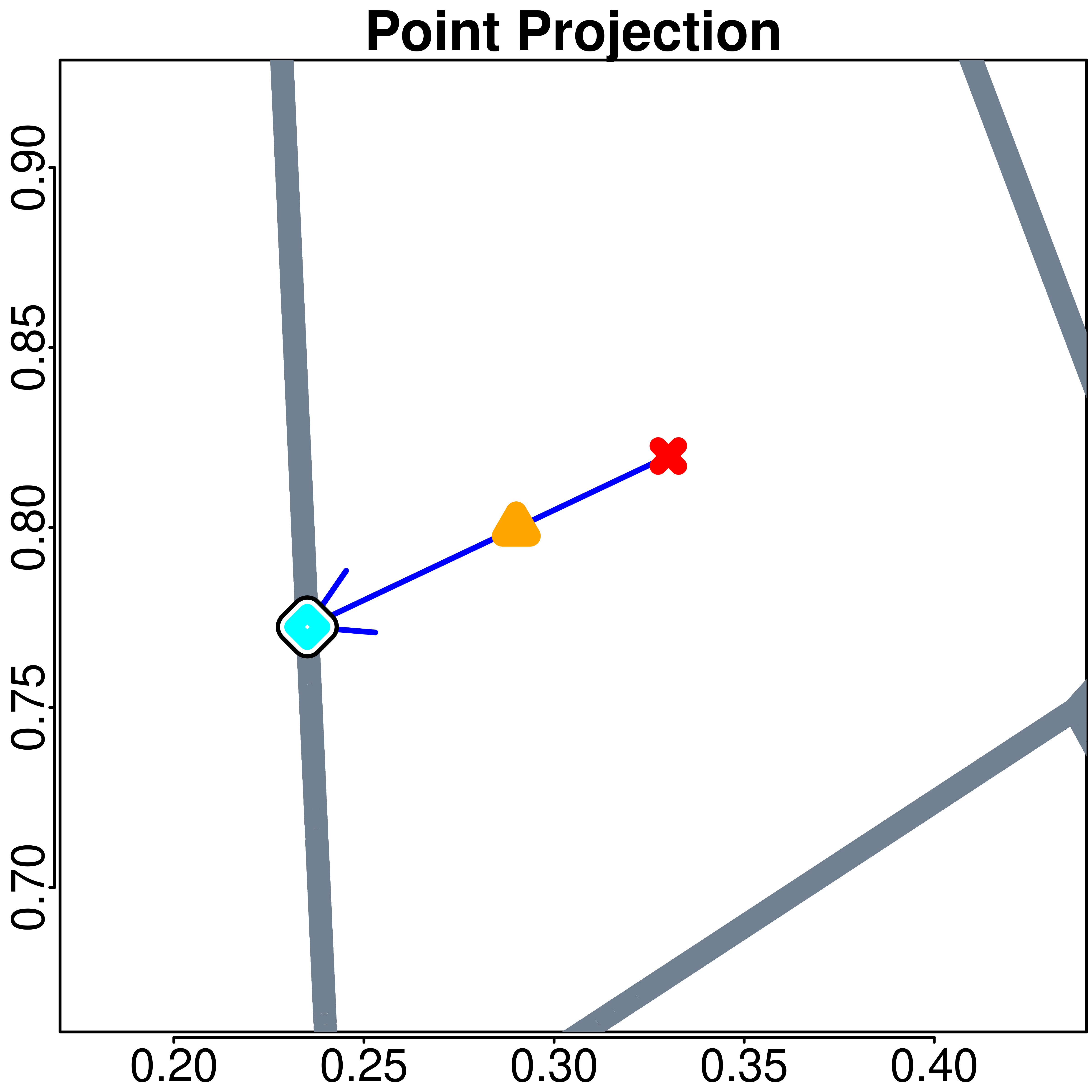}
    \includegraphics[width=0.48\textwidth]{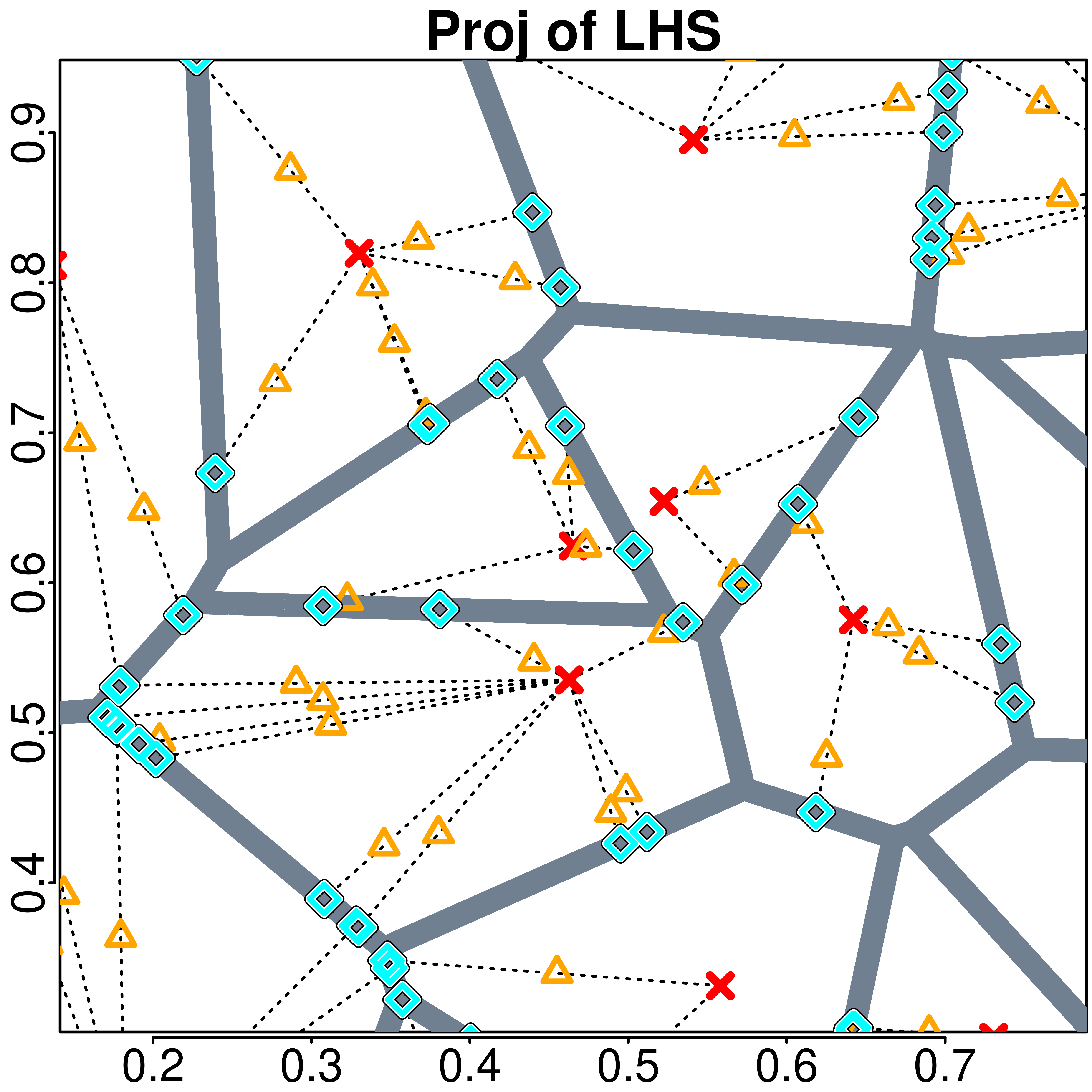}
    \caption{
    The projection of a single point (left) and an entire LHS sample (right) onto the boundary. Red ``x''s are design points, orange triangles are precandidates, and cyan diamonds are Voronoi candidates.}
    \label{fig:proj}
\end{figure}

Instead we propose the following.
First solve a nearest-neighbors problem with respect to the design in order to determine the Voronoi cell $\z$ is in; say it is in the $j^\textrm{th}$ cell.
This will serve as the sampled $n$.
Next, take the implied $\uu$ to be given by the direction we have to walk from $\x_j$ to end up at $\z$; that is, $\uu = \z-\x_j$.
Finally, conduct a Voronoi walk from $\x_j$ along $\mathbf{u}$ until we reach the boundary (see Figure \ref{fig:proj}, left) to obtain a sampled point.

It remains to specify how the $\z$ are generated.
One approach would be to generate samples by sampling $\z$ independently from some distribution, like the uniform distribution on $[0,1]^P$.
However, in this article we will instead jointly generate an entire sample of precandidates $\Xcp$ via LHS, a sampling strategy we denote as \texttt{proj}.
By starting with an LHS, which has good projection and coverage properties, and subsequently projecting it onto our Voronoi boundary, we hope to achieve a set of candidates throughout the space which nevertheless respects our existing design points.
Regardless of how the precandidates $\Xcp$ are generated, they can be converted into candidates in $\partial\Vc$ using Algorithm \ref{alg:proj}, which is illustrated in Figure \ref{fig:proj}, right.

\begin{algorithm}[h]
    \caption{Projection Vorcands}
    \label{alg:proj}
    \begin{algorithmic}[1] 
            \STATE {\bfseries Input:} $\X, \y, \Xcp$
            \FOR{$c \in \{1, \ldots, |\Xcp|\}$}
                \STATE $n[c] \gets \textrm{near\_neigh}(\X, \hat{\x}_c)$ \COMMENT{Compute Voronoi Cell}
                \STATE $\mathbf{U}[c,:] \gets \sqrt{P}\frac{\hat{\x}_c - \x_{n_c}}{\Vert\hat{\x}_c - \x_{n_c}\Vert}$ \COMMENT{Compute Search Direction}
            \ENDFOR
            \STATE {\bfseries Return:} Vorwalk($\X, \n, \U$) \COMMENT{Algorithm 2}
    \end{algorithmic}
\end{algorithm}

Though we are able to use the same kernel in terms of computation as in Algorithm \ref{alg:vorwalk}, the type of candidates produced by this procedure is conceptually distinct from the previous section.
Rather than choosing Voronoi cells by sampling design points, this allows us to choose Voronoi cells by sampling $\z\in[0,1]^P$ and then checking which cell $\z$ belongs to.
One motivation for this approach might be if we wish to sample the input space about evenly for the purposes of exploration, avoiding oversampling near existing design points while maintaining the equidistance property in our candidates.
But our algorithm accommodates precandidates motivated by any other desiderata.

\subsection{Choice of Metric and Boundary Issues}

We proposed sampling the Voronoi boundary as a means of finding candidates ``between" existing design points.
But the boundary of the Voronoi cell contains not only those points equidistant to two or more design points, but also those on the boundary of $[0,1]^P$.
Indeed, in high dimension, it is not \textit{a priori} inconceivable that most points sampled according to one of our algorithms would wind up missing any other design points and whizz straight to the boundary.
In this section, we will numerically investigate this phenomenon.

While we have so far covered how to sample from a Voronoi boundary given a tesselation, as we discussed in Section \ref{sec:bg_vor}, the definition of the tesselation relies on a dissimilarity metric.
Only by specifying both a sampling strategy and a metric have we fully defined a procedure for generating candidates.
We now perform some basic evaluations of the suitability of the $\ell_1$, $\ell_2$ and $\ell_\infty$ metrics when combined with the \texttt{unif}, \texttt{rect} and \texttt{proj} sampling strategies.

\begin{figure}[ht!]
    \centering
    \includegraphics[width=0.9\textwidth]{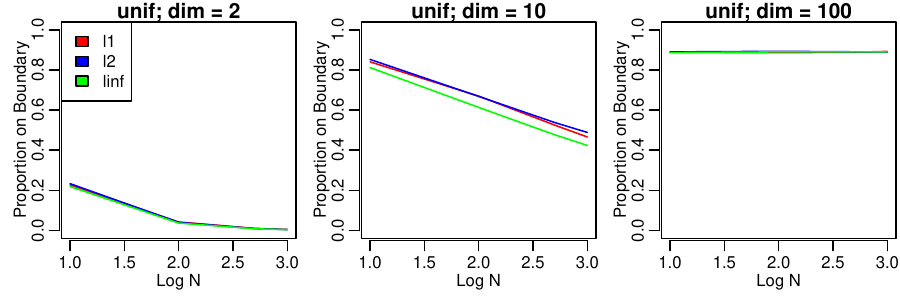}
    \includegraphics[width=0.9\textwidth]{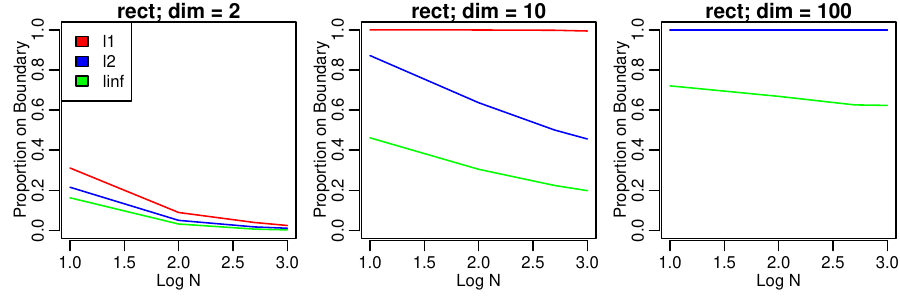}
    \includegraphics[width=0.9\textwidth]{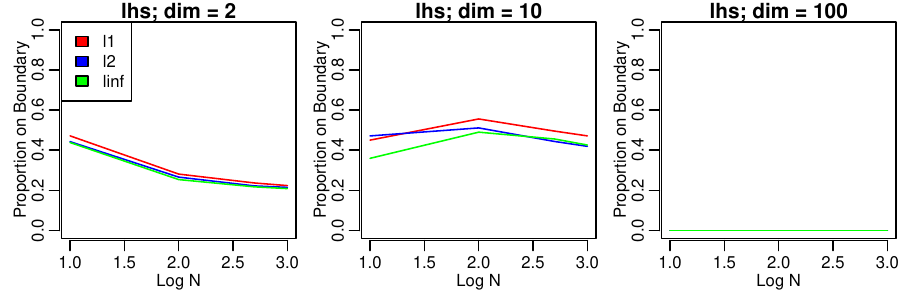}
    \caption{Propensity for vorcands to lie on $\partial[0,1]^P$ for various combinations of norms and sampling strategies.
    Each plot gives the proportion of boundary points as a function of sample sizes, and each column considers designs in a different dimension.
    \textit{Top:} Using direct sampling with a uniform measure on angles.
    \textit{Mid:} Using direct sampling of axis-aligned search directions uniformly at random.
    \textit{Bottom:} Using a projection of a random LHS.
    }
    \label{fig:edge}
\end{figure}

For sample sizes $N\in\{10,100,1000\}$, we sample designs uniformly at random for dimensions $P \in \{2, 10, 100\}$.
For each combination of metric and sampling strategy, we calculate our proposed candidates and evaluate the proportion of times that they lie in $\partial [0,1]^P$.
The results are given in Figure \ref{fig:edge}.
%
We see that, for certain settings, up to 100\% of points lie on the boundary, such that mitigating this phenomenon is a major concern. 
In general, we find that higher dimension is associated with more points on the boundaries, and higher sample sizes with fewer.
The $\ell_1$ norm in nearly all settings sends more points to the boundary than the $\ell_2$, which in turn sends more points to the boundary than the $\ell_\infty$.
This is intriguing given the diameter of the design region $[0,1]^P$ under these metrics, which is $P, \sqrt{P}$ and $1$ respectively, and therefore the fact that under the $\ell_\infty$ norm, the diameter of the unit hypercube is independent of dimension.

Comparing the two direct sampling methods \texttt{rect} and \texttt{unif}, we find that neither has an advantage in all settings. 
But when combined with the $\ell_\infty$ norm, \texttt{rect} is less likely to encounter the boundary across all test conditions than \texttt{unif} with any norm.
We find that \texttt{proj} has a non-monotone relationship between dimension and boundary prominence, and the norm used seems to have little effect.
Indeed, it is able to achieve next to no boundary points in dimension $P=100$.

\subsection{Final Implementation}\label{sec:implementation}

Our framework leaves open considerable room for specific implementation decisions which can have a significant impact on the performance of the algorithm.
In this section, we report a configuration which we found was effective after considerable trial and error.
However, there are surely other attractive possible alternatives, and the best strategy may be dependent on the problem at hand.

On the basis of the analysis of the preceding subsection, we found that \texttt{rect} combined with $\ell_\infty$ norm was the most promising direct sampling method, and that \texttt{proj} performed well in high dimension with any norm.
However, when assessing these approaches for performing BO, we found that 
on some numerical problems, \texttt{rect} did well, while on others the \texttt{proj} outperformed it. 
Combining them by alternating which one is used iteration to iteration provided a consistent algorithm.
To keep things simple, we also use the $\ell_\infty$ norm with \texttt{proj}.
We additionally considered combining the two approaches by taking half of our candidates from one scheme and half from the other.
However, we found that oftentimes one approach would dominate in terms of EI achieved, and we would in practice simply be using one of the two algorithms with half the candidate size.
This is similar to the strategy of TuRBO \cite{eriksson2019scalable}, which uses trust regions of different sizes \textit{across} acquisition iterations, rather than within.  
It is also inline with other BO works which have founding alternating acquisition schemes to perform well \cite{bull2011convergence,kim2024enhancing}.

Furthermore, when using the \texttt{unif} or \texttt{rect} strategies, we found that rather than using a completely uniform distribution on the Voronoi cells, using one which is biased in favor of the design point corresponding to the current best observed function value performed better; we made sure that at least $2P$ candidates started at the current best observed optimum and sampled the remaining candidates uniformly at random from the remaining Voronoi cells.
One could imagine a more sophisticated strategy which considers, e.g., the ranks of the design points.


Finally, we found significant advantages in the simulation studies when going only halfway to a boundary from the design point when a candidate was supposed to lie on the boundary $\partial [0,1]^P$.  We have implemented this strategy for all numerical results thus far in the paper. 
For an illustration, see the bottom right of Figure \ref{fig:cands_compare_low} where no Voronoi points were proposed exactly on the boundary.  

\section{Numerical Experiments}\label{sec:exp}

We consider 
eight deterministic test problems in ten or more dimensions.
We use a GP with a Gaussian kernel and EI acquisitions throughout, varying only the manner by which the EI is optimized.
Hyperparameters are set via maximum likelihood for each of the first 200 iterations, and then once every 25 iterations thereafter, with L-BFGS-B initialized on the previous iteration's optimum using the R package \texttt{laGP} \citep{laGP}.
To compare to the standard continuous inner-optimization-based acquisition, we implement multistart L-BFGS-B (\texttt{opt}) with initializations sampled from a random LHS of size $2P$ as well as one initialization made at the current best observed point.

However, we view our primary basis of comparison as being existing candidate schemes. 
We include a Sobol Sequence (\texttt{sobol}), a random LHS (\texttt{lhs}), and Tricands (\texttt{tri}).
With Tricands, we quickly ran into its computational limits.
On our ten dimensional test problems, running Tricands for only 100 blackbox evaluations cost as much as running \texttt{opt} for the entire duration (500-800 evaluations).
We were not able to run it even for 100 evaluations on the 12 dimensional (Lunar) test problem, nor any higher dimensional ones.
The method developed in this article is denoted \texttt{vor}, and is implemented as described in Section \ref{sec:implementation}.
Finally, we include the classical Nelder Mead method (\texttt{NM}) and L-BFGS-B with finite difference gradients applied directly to the black-box  (\texttt{BFGS}).

Throughout we track the best observed value $y_{\min}$ over the course of acquisitions for increasing budgets of black-box evaluations over the course of an optimization run. 
We additionally measure computational expense using elapsed real time for GP-based methods, which includes hyperparameter fitting, acquisition sub-problem solving, and black-box evaluation.
We did 100  repetitions of each experiment, using an initial design of size $3P$ sampled from a random LHS; the initial design is shared by all methods.
All candidate-based approaches used $|\mathcal{X}_C| = \min(5000,100P)$
following \citet{eriksson2019scalable}.
The experiments were conducted on a heterogeneous computing cluster; each node ran all methods so as to make execution time comparisons valid.
Code reproducing these results is available in a public git repository.\footnote{
\ifarxiv
    \url{https://github.com/NathanWycoff/vorcands}
\fi
}

Our primary figures illustrating aggregated results are given as Figures \ref{fig:toy}, \ref{fig:game} and \ref{fig:eco} and are discussed in the following three subsections.
Figures \ref{fig:toy_box}, \ref{fig:game_box} and \ref{fig:eco_box} show the full distribution of best observed function values, illustrating run-to-run variation for all test problems.

\subsection{Toy Problems}\label{sec:exp_toy}

\begin{figure}[ht!]
    \centering
    \includegraphics[width=0.9\textwidth,trim={5em 7em 16.6em 20.2em},clip]{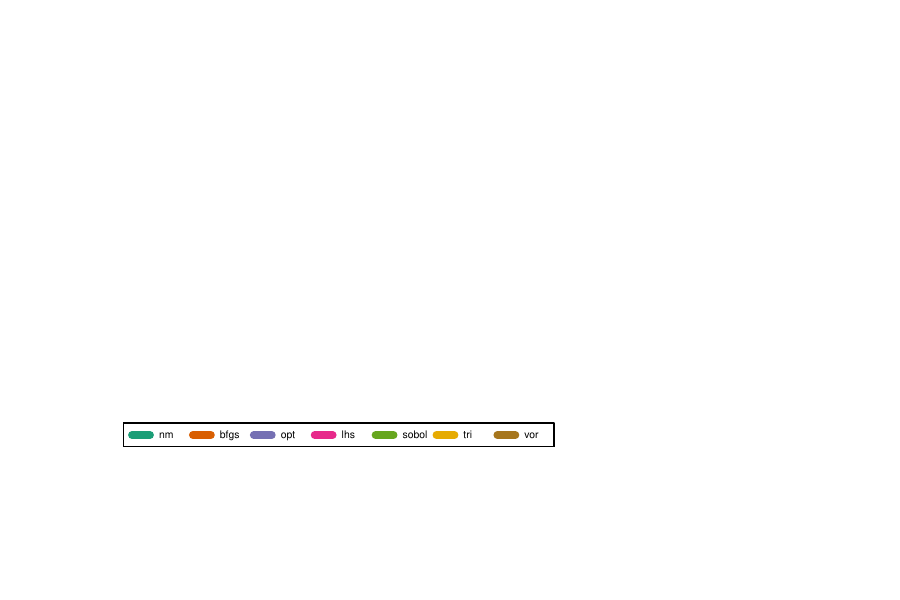}
    
    \includegraphics[width=0.42\textwidth,trim=0 20 0 0, clip=TRUE]{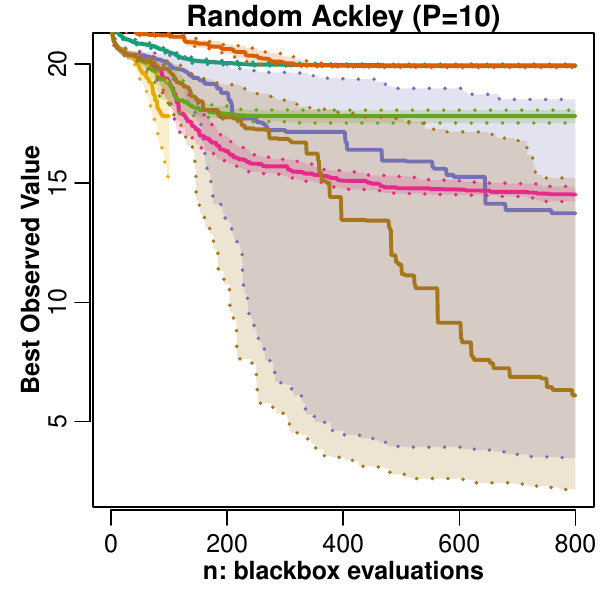}
    \includegraphics[width=0.42\textwidth,trim=0 20 0 0]{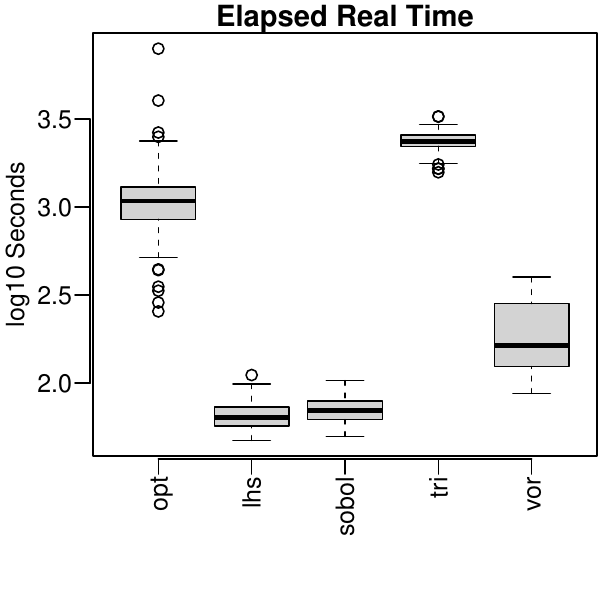}    

    \includegraphics[width=0.42\textwidth,trim=0 20 0 0, clip=TRUE]{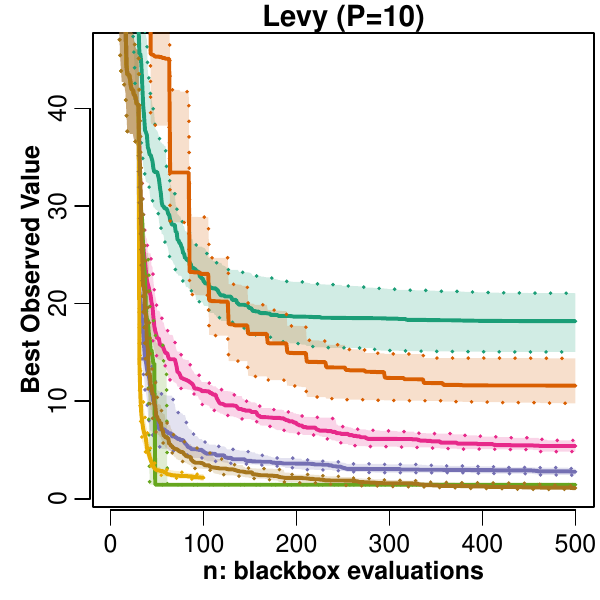}
    \includegraphics[width=0.42\textwidth,trim=0 20 0 0]{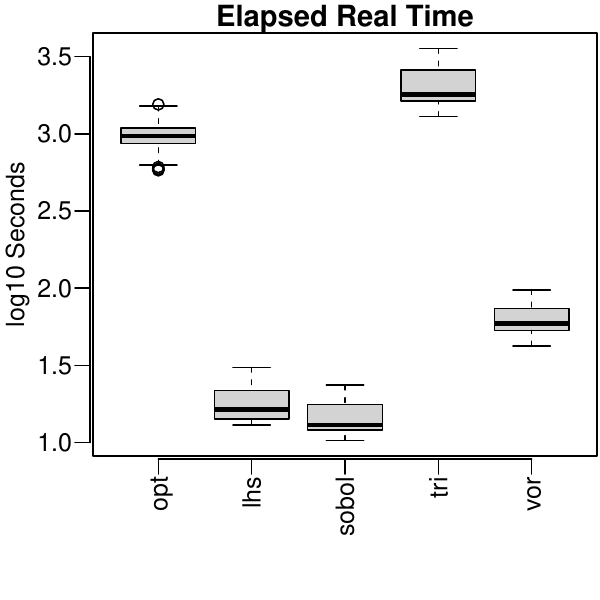}
    
    \includegraphics[width=0.42\textwidth]{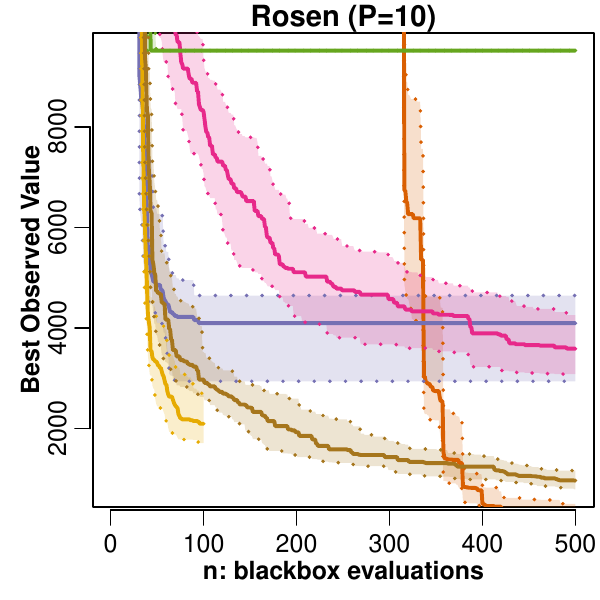}
    \includegraphics[width=0.42\textwidth]{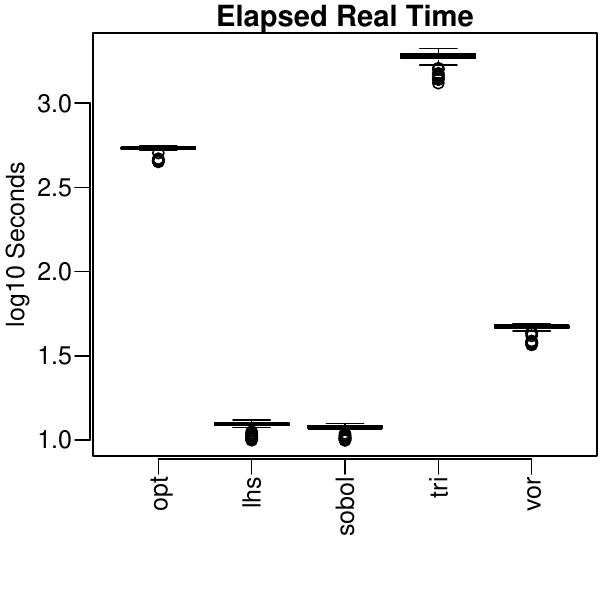}

    \caption{Performance on toy problems, one in each row.
    Line charts give best observed function value as a function of design size. 
    The distribution of performances over repetitions is given in the first two boxplots.
    Rightmost boxplots summarize cumulative execution time. \label{fig:toy}}
\end{figure}

We first consider three popular test functions in dimension 10: the Ackley, Levy, and Rosenbrock functions (Figure \ref{fig:toy}).
We use a randomized version of the Ackley function with optimum placed uniformly in $[0,1]^P$ via translation by a uniform random vector.
Otherwise, we use the standard configurations as specified in the Virtual Library of Simulation Experiments.\footnote{\url{https://www.sfu.ca/~ssurjano/index.html}} 
The first column indicates the distribution of progress on 
best observed value over every iteration of acquisition in terms of median and 90\% error-bars calculated from
the 100 Monte Carlo replicates.  
The second column provides a summary of execution time for GP-based methods, cumulatively over all acquisitions. 

\begin{figure}[ht!]
    \centering
    \includegraphics[width=0.32\textwidth]{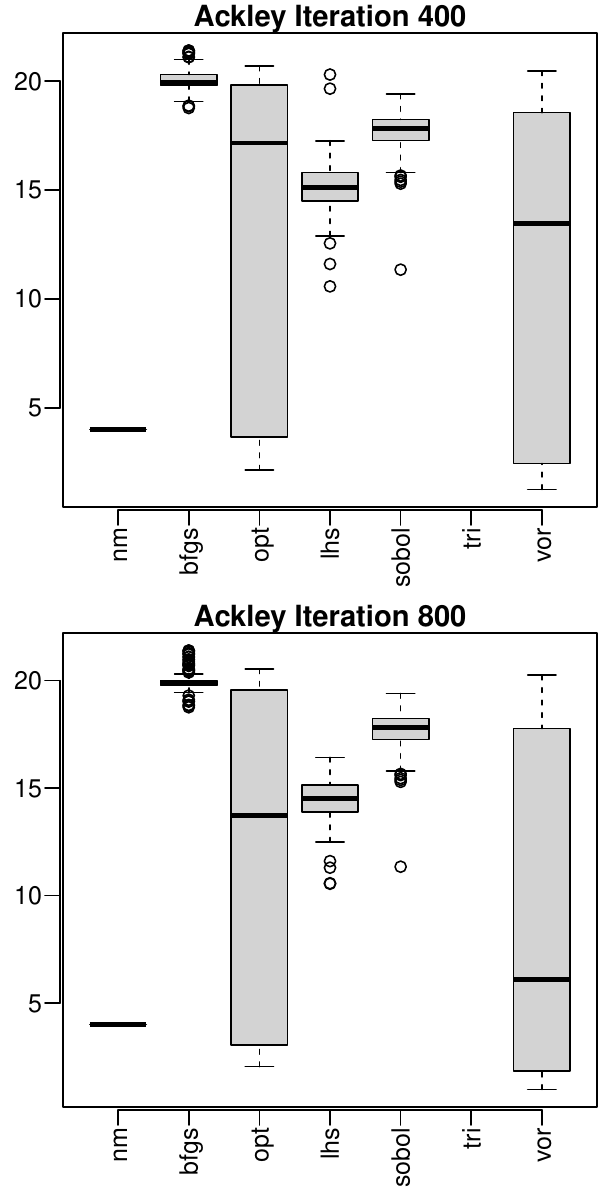}
    \includegraphics[width=0.32\textwidth]{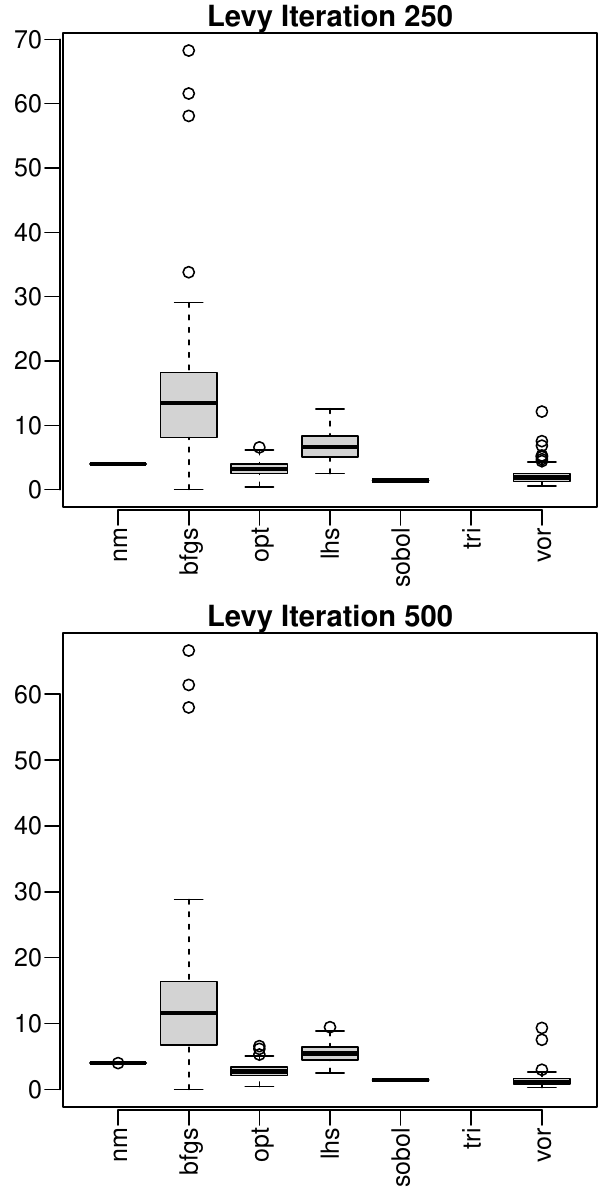}
    \includegraphics[width=0.32\textwidth]{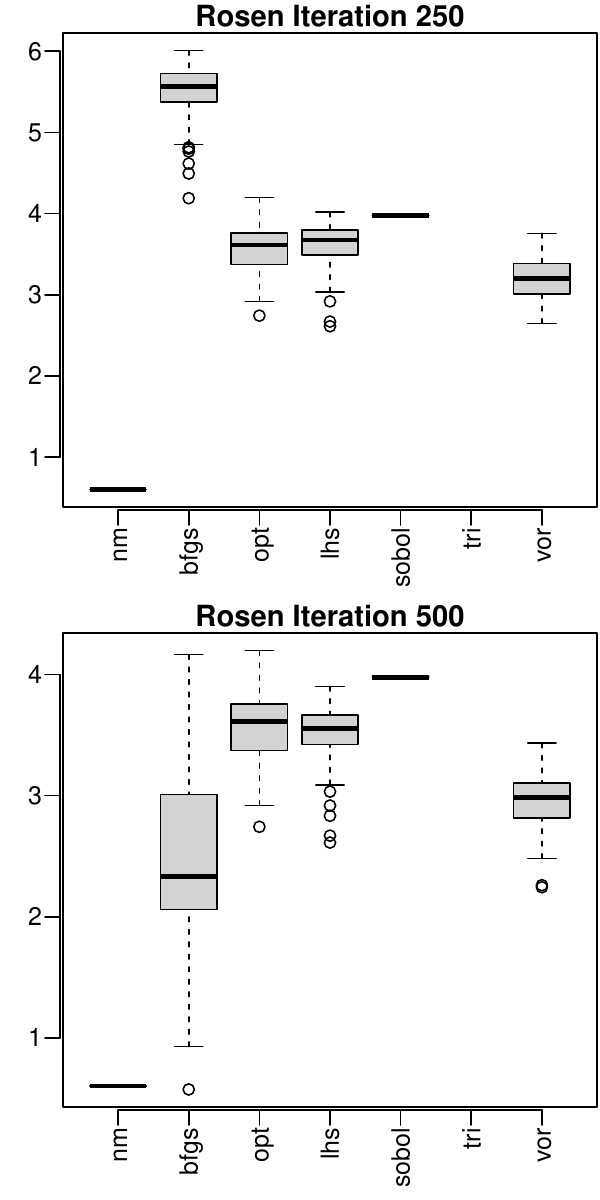}
    
    \caption{Boxplots giving the entire distribution of best observed function values for all 100 Monte Carlo replicates. 
    Each pair of plots corresponds to a test problem, with the top plot showing midway through the budget and the bottom plot showing the final budget.
    }
    \label{fig:toy_box}
\end{figure}

On the Ackley function, the \texttt{vor} method is able to make substantial progress in the median run.
In its best runs, \texttt{opt} gets pretty close, but the other candidate methods fare far worse.
Figure \ref{fig:toy_box} shows there is nevertheless considerable variation in solution quality for these two methods as the initial design is varied.
The Tricands comparator starts off well, but we were unable to continue beyond acquisition 100 due to the extreme computational expense of acquisition.
Observe in the top-right panel that it is the slowest of all methods on average even though it was stopped after only 100 evaluations.
This is also the case for the other test cases, in rows two and three.

On Levy, in the second row of Figure \ref{fig:toy}, we find that the \texttt{sobol} method actually does quite well.
This may be because the optimum is in the exact center of the space, and the Sobol sequence, by its recursive refinement of the space, happens to propose this exact point.
Again, \texttt{vor} seems to edge out a win in terms of best observed function value relative to \texttt{opt} while \texttt{lhs} lags behind.

On Rosenbrock, the \texttt{sobol} competitor is unable to make much progress, while this time \texttt{vor} is actually able to build a substantial lead over \texttt{opt}, which seems to get stuck.
Intriguingly, BFGS manages to find superior solutions near the end of the study on this example.
As \citet{kok2009locating} discuss, the Rosenbrock function has many stationary points in high dimension (we use what those authors describe as ``Variant B"), and BFGS can be expected to both deal with the ill-conditioning of the problem as well as escape saddlepoints, possibly explaining its average-case success. 
However, the worst BFGS initializations do worse than the worst \texttt{vor} runs (see Figure \ref{fig:toy_box}), potentially owing to convergence to local optima.

On all problems, Vorcands has execution time more similar to LHS candidates than to continuous, numerically optimized acquisition. In general, we find that \texttt{vor} is able to obtain even better median results than \texttt{opt} in significantly less time on these problems.
Our \texttt{vor} competitor also does much better than \texttt{lhs} and is more consistent than \texttt{sobol} with only slightly more time.

\subsection{Game Problems}\label{sec:exp_game}

We next investigate three test problems that arise from automatic solving of simple video game interfaces: the Lunar landing ($P=12$), Push ($P=14$) and Rover ($P=60$) problems.  
The Lunar Landing problem \cite{openaigym} involves configuring a spaceship to land at a particular point on a 2d lunar surface in two dimensions and without much downwards velocity in order to obtain a good ``score'', i.e., $y$-value.
The Push problem \cite{wang2017batched} involves two robot hands pushing a box into a designated area, with the parameters controlling the dynamics of the robot arms, and score measured by the distance between the achieved and desired location of the box.
The Rover problem \cite{wang2018batched} consists of setting the trajectory of a rover navigating an environment.
We modified these three problems to be deterministic by changing the standard deviation of any normal variates to zero.
On these problems, we were not able to run Tricands even for the initial design.

\begin{figure}[ht!]
    \centering

    \includegraphics[width=0.9\textwidth,trim={5em 7em 16.6em 20.2em},clip]{images/legend.pdf}
    
    \includegraphics[width=0.42\textwidth,trim=0 20 0 0,clip=TRUE]{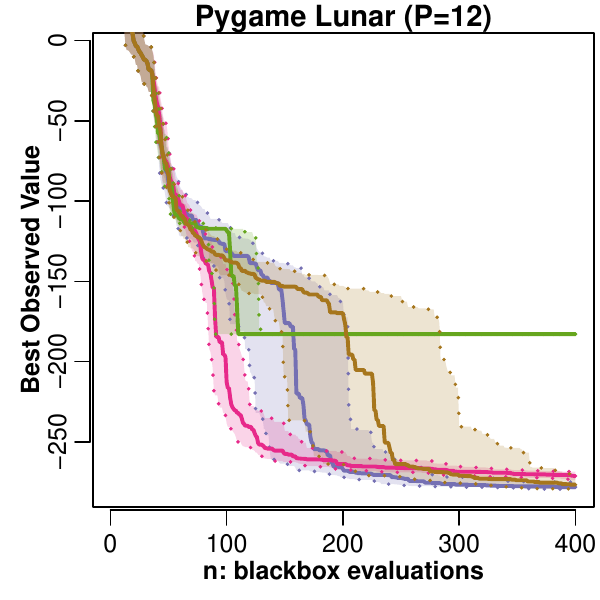}
    \includegraphics[width=0.42\textwidth,trim=0 20 0 0]{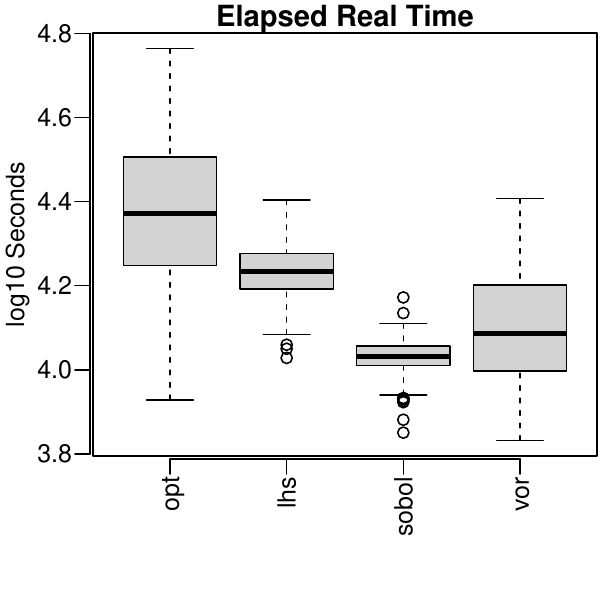}
    
    \includegraphics[width=0.42\textwidth,trim=0 20 0 0,clip=TRUE]{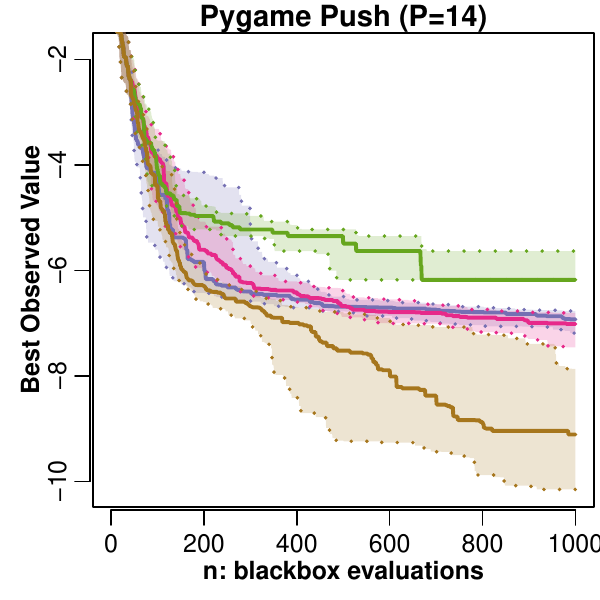}
    \includegraphics[width=0.42\textwidth,trim=0 20 0 0]{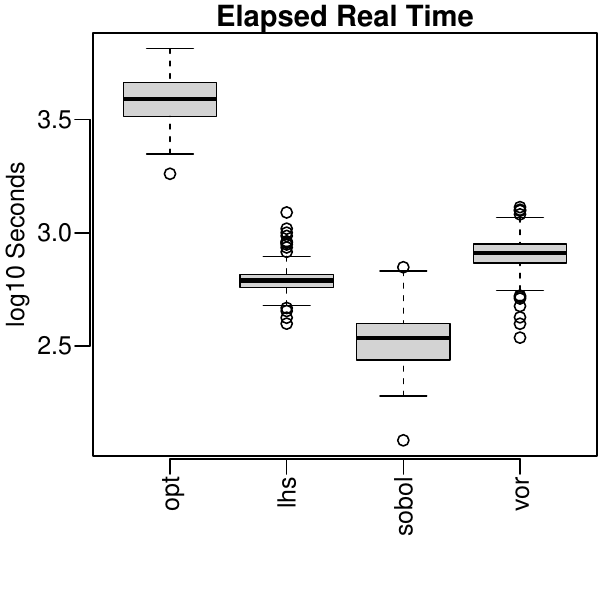}
    
    \includegraphics[width=0.42\textwidth]{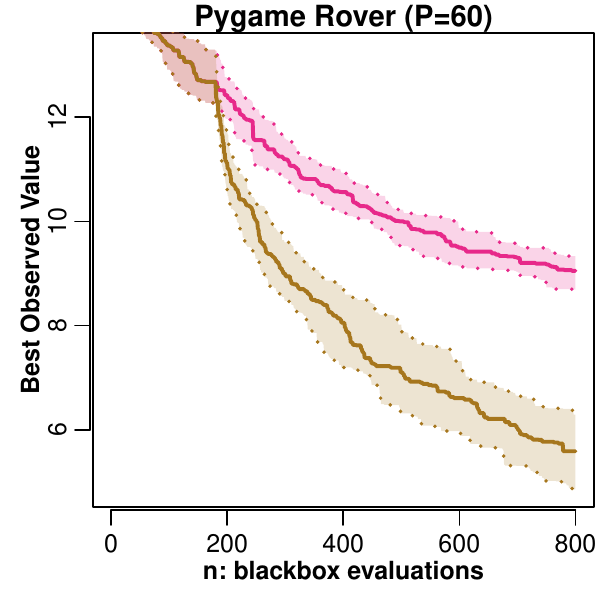}
    \includegraphics[width=0.42\textwidth]{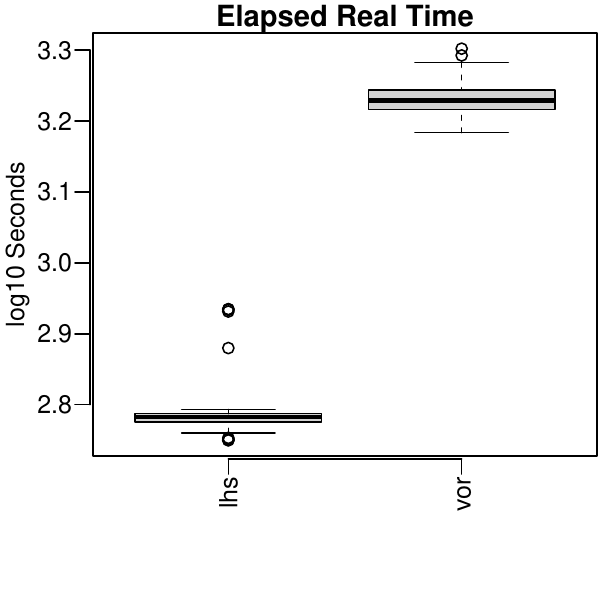}
    
    \caption{Performance on game problems; see Figure \ref{fig:toy} caption.}
    \label{fig:game}
\end{figure}

\begin{figure}[ht!]
    \centering
    \includegraphics[width=0.32\linewidth]{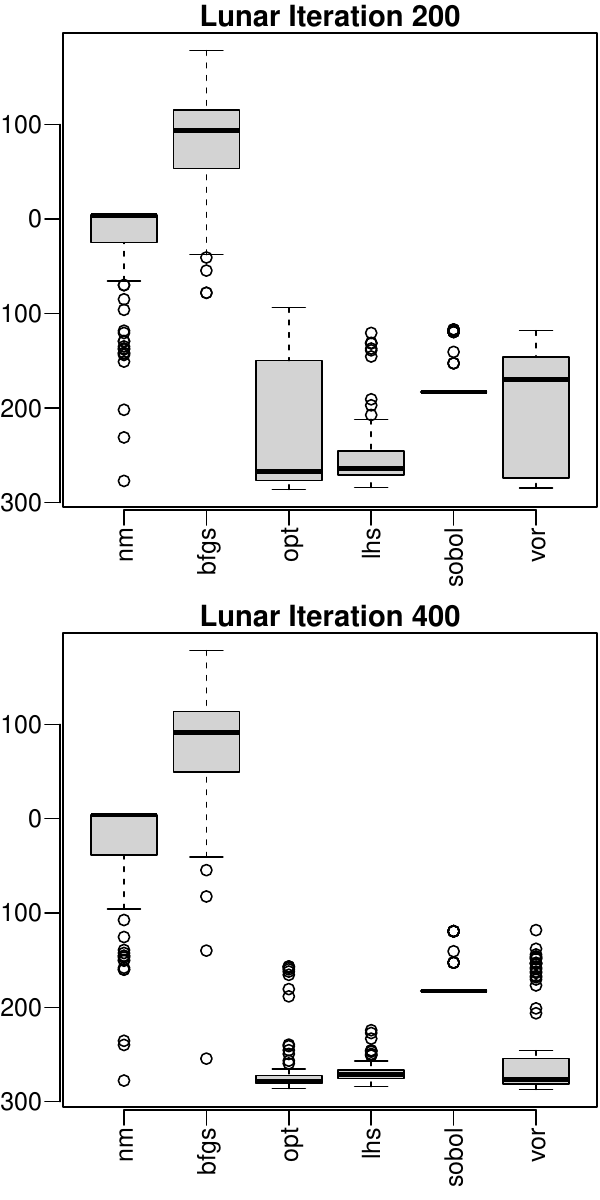}
    \includegraphics[width=0.32\linewidth]{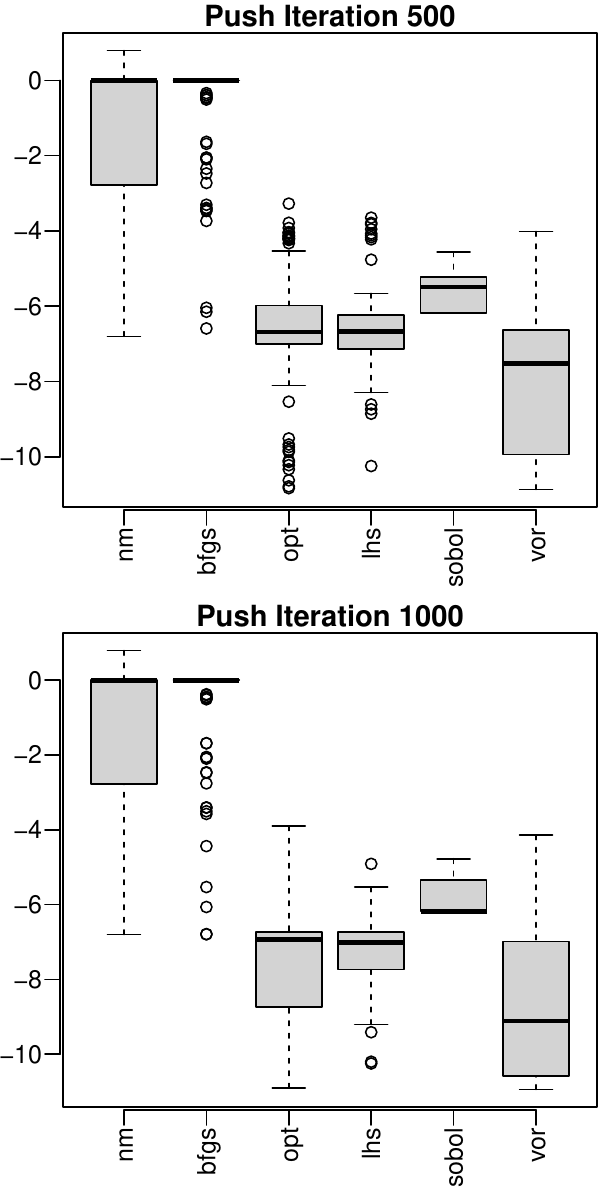}
    \includegraphics[width=0.32\linewidth]{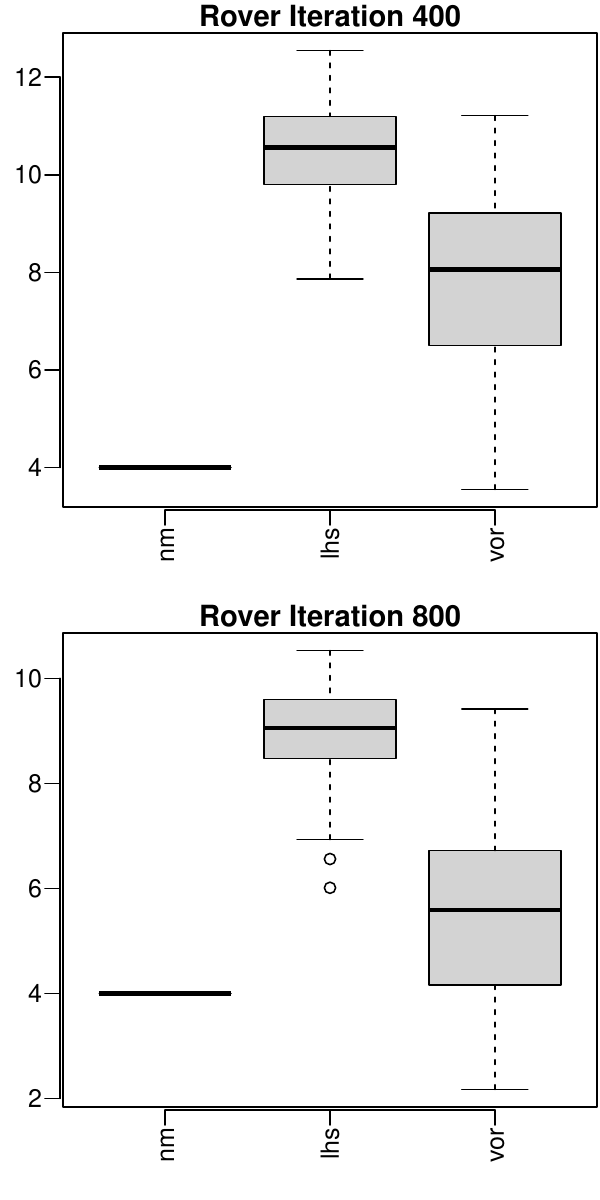}
    \caption{Distribution of performance on game problems; see Figure \ref{fig:toy_box} caption.}
    \label{fig:game_box}
\end{figure}

See Figure \ref{fig:game}, summarizing the results of our benchmarking on these test problems, which is laid out similarly to Figure \ref{fig:toy}.
On the Lunar problem, the \texttt{lhs} approach actually takes an early lead.
Upon investigation, we found that this was because it is possible to achieve a $y_{\min}$ of around -250 by simply setting the 11th input variable to a value very near 1, regardless of the other values. 
Since an LHS has very good projection properties, it is able to find this earlier than any other method, and indeed slightly ahead of \texttt{opt}, the second best competitor.
This is the only problem where \texttt{vor} lags significantly  behind \texttt{opt}, and we suspect that \texttt{vor} is being hindered by our approach of going only halfway to the boundary (Section \ref{sec:implementation}).
Although the median for \texttt{opt} is indeed lower than that of \texttt{vor}, the interquartile ranges essentially coincide; see Figure \ref{fig:game_box}.
On the Push problem, \texttt{vor} does significantly better than \texttt{opt}, which is similar to the \texttt{lhs} method.
The execution time for \texttt{vor} is about the same as for \texttt{lhs}. 
For the Rover, we found that the simulator would crash when duplicate inputs were provided, which was attempted by the \texttt{opt}, \texttt{bfgs} and \texttt{sobol} methods when they proposed points on the boundary of the space. 
We find that \texttt{vor} beats \texttt{lhs} in terms of best-observed function value, though note that \texttt{nm} consistently outperforms both (Figure \ref{fig:game_box}). 

\subsection{Ecological test problems}

We conclude our empirical work with a benchmarking exercise derived from two test problems from ecology, both involving minimizing
the negative log marginal likelihood depicting the fit of a statistical model to data.
One is a 10 dimensional forest ecosystem model \cite{smith2023}, the other a 24 dimensional cholera model \cite{IF2}; in both
cases the marginal likelihood is approximated via a particle filter.

The first problem is a stochastic differential equation model of cholera dynamics in the Dhaka district of Bangladesh, taken from \citet{king_ionides_dhaka}.
The resulting so-called ``Dhaka function'' has previously been used as a test problem for methodological developments that use iterated filtering \cite{ionides2011_if, IF1, IF2}, a procedure specific to this class of problems.
When global optimization is of primary interest, iterated filtering approaches suggest the use of multiple searches from randomized starting values \cite{IF2}.
This requires a much larger number of function evaluations than the use of a single-run global optimization algorithm (details below). 
Here we investigate whether our Voronoi candidate methodology can find global optima that are comparable to those reported in \citet{king_ionides_dhaka}, while using a single-run global optimization strategy instead of multiple searches, as well as over 10 times fewer function evaluations overall. 

The second problem is a partially observed Markov process that uses a two-dimensional discrete dynamical system to describe the interaction between foliage and labile carbon in a forest ecosystem, taken from \citet{smith2023}. This dynamical system requires the optimization of ten free parameters, and will be referred to hereafter as ``pomp10''. In addition to this, the objective function for pomp10 is bimodal, having a large plateau of high density, but also a narrow mode with higher density. \citet{smith2023} estimate parameters for pomp10 using particle Markov Chain Monte Carlo \cite[pMCMC;][]{pMCMCAndrieu}. Given the relative inefficiency of pMCMC compared to alternatives \cite{bhadra2010}, we aim to explore whether our Vorcands-based-BO able to find the higher density mode of the objective function in considerably fewer function evaluations. Independent of the performance of the global optimization, locating this second mode is valuable for finding an area of high posterior density to start pMCMC chains in applications where uncertainty quantification is of importance \cite{IF2}.

\begin{figure}[ht!]
    \centering

    \includegraphics[width=0.9\textwidth,trim={5em 7em 16.6em 20.2em},clip]{images/legend.pdf}
    
    \includegraphics[width=0.45\textwidth,trim=0 20 0 0,clip=TRUE]{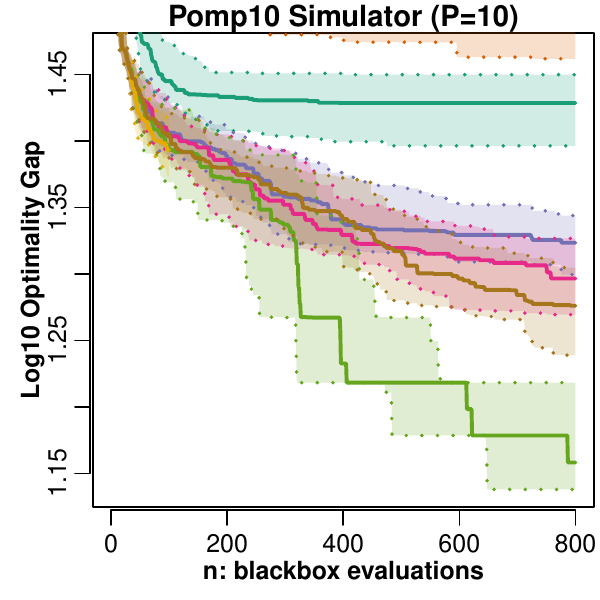}
    \includegraphics[width=0.45\textwidth,trim=0 20 0 0 ]{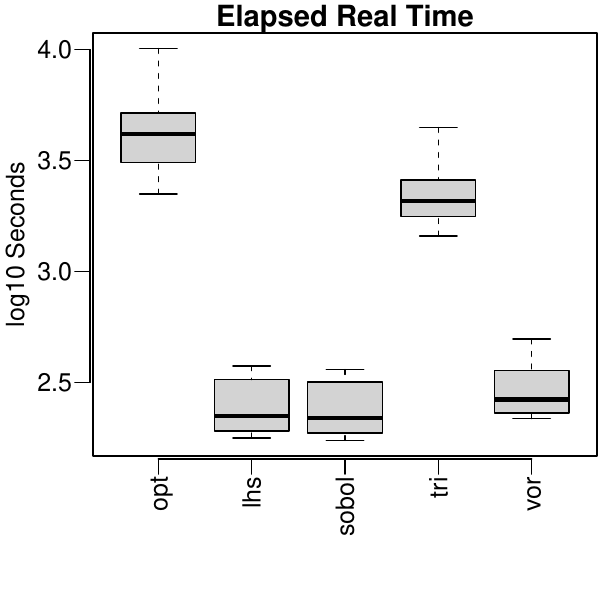}

    \includegraphics[width=0.45\textwidth]{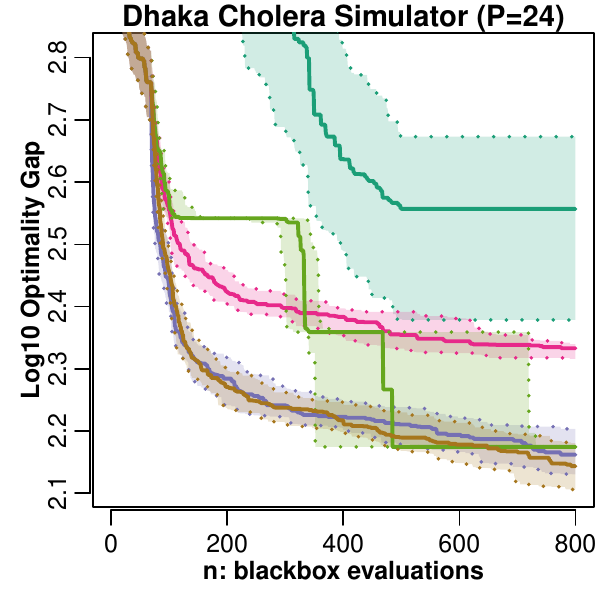}
    \includegraphics[width=0.45\textwidth]{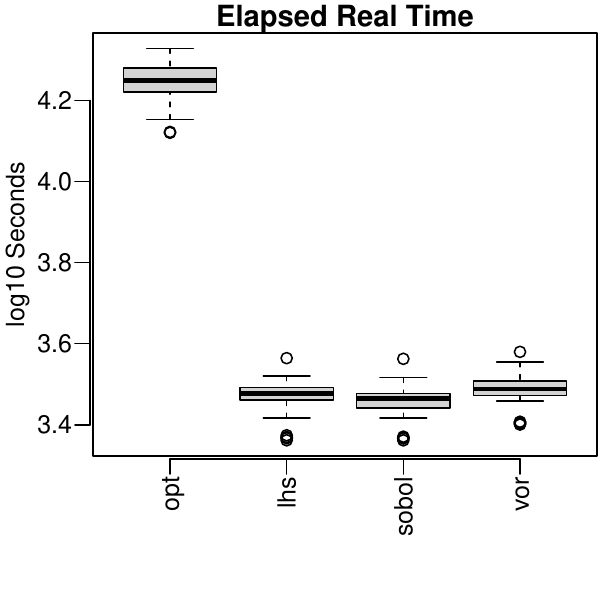}
    
    \caption{Performance on ecology test problems; see Figure \ref{fig:toy} caption.}
    \label{fig:eco}
    
\end{figure}

 As a slight change to the display of results from the previous two figures, here the first column of Figure \ref{fig:eco} shows, on the $y$-axis, the gap between the current best observed function value and the best value achieved by the more computationally demanding traditional methods which are far more profligate with their evaluations.   The optimal reference points for the Dhaka function represent the maximum likelihood value reported by \citet{king_ionides_dhaka} using a maximization by iterated filtering algorithm \cite{IF1}.   The optimal points for the pomp10 function represent the maximum likelihood value reported by \citet{smith2023}, obtained from 100,000 pMCMC iterations.

On pomp10, we find that \texttt{vor} is able to retain a favorable rate of improvement as the experiment goes on compared to \texttt{opt} and \texttt{lhs}.
We are for this problem able to run Tricands for 100 iterations, but this is not sufficient for us to make good progress.
Intriguingly, \texttt{sobol} performs quite well on this problem, though we note that the single best observed setting was found by \texttt{nm} (though \texttt{nm} has a median performance inferior to other strategies).
On the 24 dimensional Dacca problem, \texttt{vor} and \texttt{opt} are in lockstep, and lead the other methods for most evaluation budgets, though for budgets of about 550-700 \texttt{sobol} again offers good results. 
\begin{figure}[ht!]
    \centering
    \includegraphics[width=0.32\linewidth]{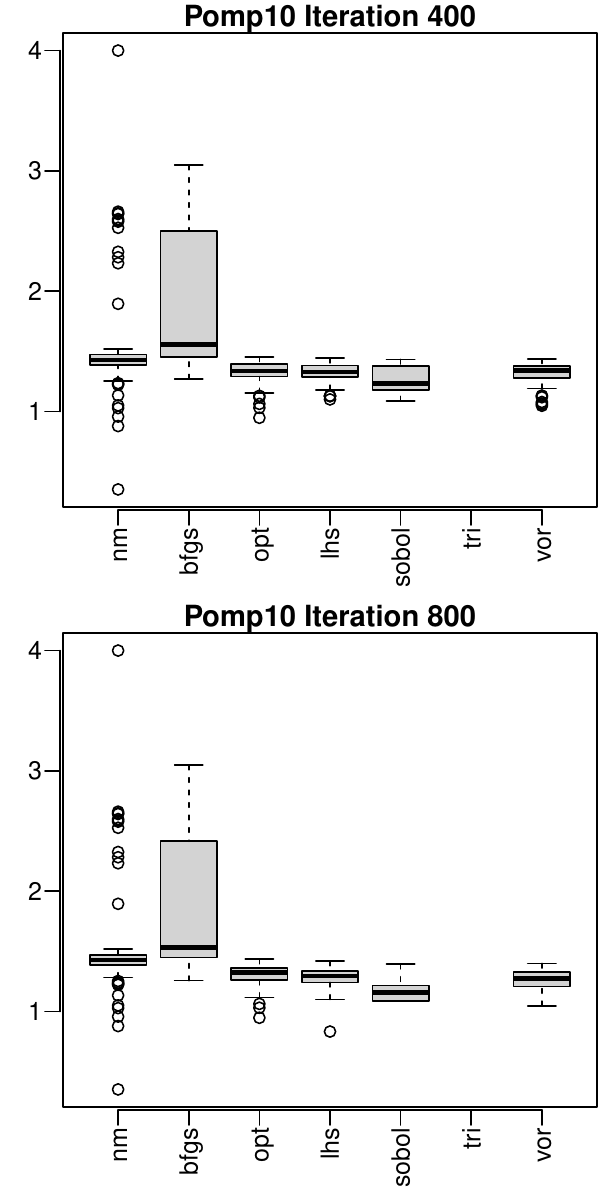}
    \includegraphics[width=0.32\linewidth]{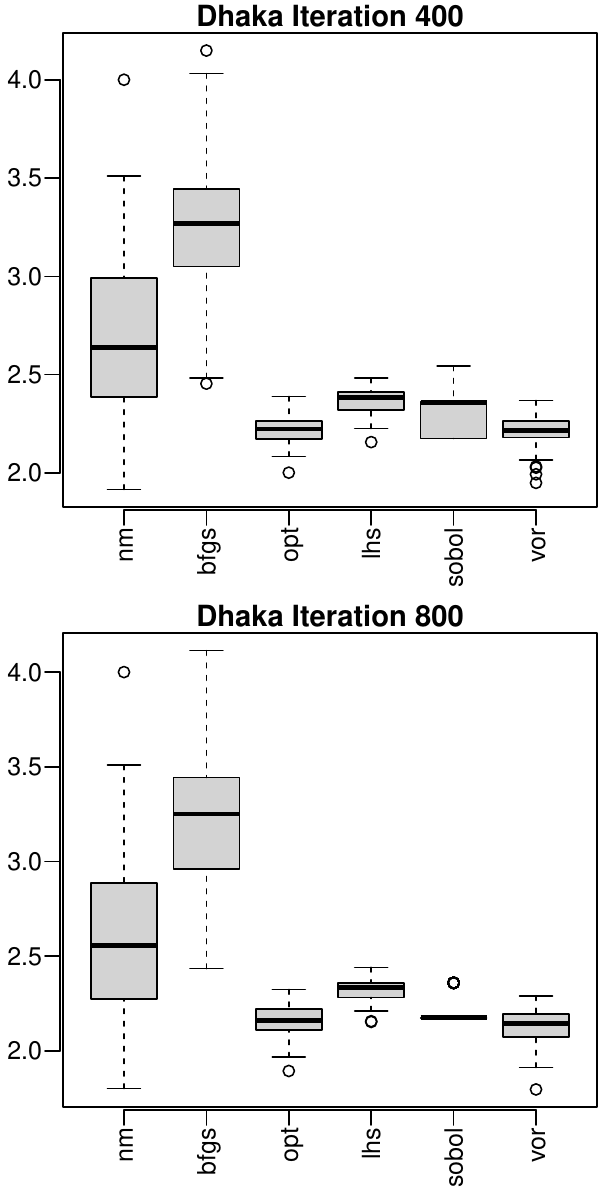}

    \caption{Distribution of performance on ecology problems; see Figure \ref{fig:toy_box} caption.}
    \label{fig:eco_box}
\end{figure}
However, Figure \ref{fig:eco_box} indicates that \texttt{sobol} seems to be consistently achieving a particular strategy by the end of the trial, as there is essentially no run-to-run variation. 
In our view, this suggests that if the budget were extended it would probably not continue to improve. 
This demonstrates the advantage of the adaptive nature of \texttt{vor} over \texttt{sobol}, which essentially has a fixed resolution.
We see that the advantage offered by the alternating Voronoi strategy over a simple random LHS is substantial in terms of best observed function value while costing about the same amount of computations, as on this nontrivial 24 dimensional function candidate generation time is dwarfed by hyperparameter fitting and function evaluation costs.
The \texttt{opt} method takes significantly longer than the other methods on both problems.

\section{Discussion}\label{sec:disc}

We proposed Vorcands, a metaheuristic for generating candidates for sequential black-box problems based on acquisition functions with a focus on EI/GP-based BO.
In that context, we found that this approach was superior to gradient-based optimization on 7/8 test problems in dimensions ranging from 10 to 60 in terms of best observed function value and dominated it on all problems in terms of execution time.
It also dominated using an LHS as a candidate set in terms of best observed function value while being comparable in execution time.
Compared to the Sobol sequence, its performance was more consistent.
And while Tricands appeared somewhat promising to the extent that it could be used, computational considerations essentially ruled out its use on our problems which had moderate budget and dimension.
We conclude that the use of candidates, such as Vorcands, can be a promising alternative to continuous optimization even on high dimensional problems.

We have proposed a specific scheme for searching the Voronoi boundary.
But there are many conceivable techniques for sampling such points, and we cannot rule out that some are superior to those investigated here.
Furthermore, we investigated only three metrics defining Voronoi cells. 
However, data-adaptive metrics, such as that metric implicitly used by the GP kernel with optimized hyperparameters, may well be more successful, if less portable to other surrogate models.
And while we only investigated GPs in this article, the methodology we propose is applicable much more broadly.
Indeed, the use of candidate points is more flexible than BFGS-based optimization of EI, since we do not need to evaluate gradients of the acquisition.

This opens the door to integration with more complicated surrogates, such as Bayesian deep GPs employing Markov Chain Monte Carlo \citep{sauer2023active}.
Furthermore, while we only investigated EI in this article, we expect this method to perform well for any acquisition function which tends to increase when moving away from design points: any acquisition function which searches ``between" them. 
Additionally, we considered only single objective, deterministic, unconstrained optimization in this article. 
But there are no clear obstacles to applying Vorcands to multi-objective, stochastic or constrained acquisition problems merely by switching out the acquisition function. 
With some adaptation, a similar idea could also be profitable in the context of mixed integer problems or other discrete search spaces.
However, one problem domain to which Vorcands does not straightforwardly apply is that of batch acquisition: in such a case, we may not wish for candidate points to be located halfway between design points if several acquisitions are to be made in the same part of the space.
Future work should investigate alternative strategies.

\section*{Data Availability Statement}

Code reproducing all figures in this article, including code implementing the studied computer simulators, is available at \url{https://github.com/NathanWycoff/vorcands}. 
This article is not associated with any datasets \textit{per se}.

\section*{Acknowledgments}

Author NW gratefully acknowledges funding from the Massive Data Institute and the McCourt Institute as well as support from the NSF and NGA under award \#2428033. 
Author JWS gratefully acknowledges support from the NSF AI Institute in Dynamic Systems under award NSF \#2112085. Author ASB is grateful for NSF support through award \#2436164.  Author
RBG is grateful for NSF support under awards \#2152679 and \#2318861.

\bibliography{sn-bibliography}


\begin{thebibliography}{49}
\ifx \bisbn   \undefined \def \bisbn  #1{ISBN #1}\fi
\ifx \binits  \undefined \def \binits#1{#1}\fi
\ifx \bauthor  \undefined \def \bauthor#1{#1}\fi
\ifx \batitle  \undefined \def \batitle#1{#1}\fi
\ifx \bjtitle  \undefined \def \bjtitle#1{#1}\fi
\ifx \bvolume  \undefined \def \bvolume#1{\textbf{#1}}\fi
\ifx \byear  \undefined \def \byear#1{#1}\fi
\ifx \bissue  \undefined \def \bissue#1{#1}\fi
\ifx \bfpage  \undefined \def \bfpage#1{#1}\fi
\ifx \blpage  \undefined \def \blpage #1{#1}\fi
\ifx \burl  \undefined \def \burl#1{\textsf{#1}}\fi
\ifx \doiurl  \undefined \def \doiurl#1{\url{https://doi.org/#1}}\fi
\ifx \betal  \undefined \def \betal{\textit{et al.}}\fi
\ifx \binstitute  \undefined \def \binstitute#1{#1}\fi
\ifx \binstitutionaled  \undefined \def \binstitutionaled#1{#1}\fi
\ifx \bctitle  \undefined \def \bctitle#1{#1}\fi
\ifx \beditor  \undefined \def \beditor#1{#1}\fi
\ifx \bpublisher  \undefined \def \bpublisher#1{#1}\fi
\ifx \bbtitle  \undefined \def \bbtitle#1{#1}\fi
\ifx \bedition  \undefined \def \bedition#1{#1}\fi
\ifx \bseriesno  \undefined \def \bseriesno#1{#1}\fi
\ifx \blocation  \undefined \def \blocation#1{#1}\fi
\ifx \bsertitle  \undefined \def \bsertitle#1{#1}\fi
\ifx \bsnm \undefined \def \bsnm#1{#1}\fi
\ifx \bsuffix \undefined \def \bsuffix#1{#1}\fi
\ifx \bparticle \undefined \def \bparticle#1{#1}\fi
\ifx \barticle \undefined \def \barticle#1{#1}\fi
\bibcommenthead
\ifx \bconfdate \undefined \def \bconfdate #1{#1}\fi
\ifx \botherref \undefined \def \botherref #1{#1}\fi
\ifx \url \undefined \def \url#1{\textsf{#1}}\fi
\ifx \bchapter \undefined \def \bchapter#1{#1}\fi
\ifx \bbook \undefined \def \bbook#1{#1}\fi
\ifx \bcomment \undefined \def \bcomment#1{#1}\fi
\ifx \oauthor \undefined \def \oauthor#1{#1}\fi
\ifx \citeauthoryear \undefined \def \citeauthoryear#1{#1}\fi
\ifx \endbibitem  \undefined \def \endbibitem {}\fi
\ifx \bconflocation  \undefined \def \bconflocation#1{#1}\fi
\ifx \arxivurl  \undefined \def \arxivurl#1{\textsf{#1}}\fi
\csname PreBibitemsHook\endcsname

\bibitem[\protect\citeauthoryear{Frazier}{2018}]{frazier2018tutorial}
\begin{botherref}
\oauthor{\bsnm{Frazier}, \binits{P.I.}}:
A tutorial on {B}ayesian optimization.
arXiv preprint arXiv:1807.02811
(2018)
\end{botherref}
\endbibitem

\bibitem[\protect\citeauthoryear{Gramacy}{2020}]{gramacy2020surrogates}
\begin{bbook}
\bauthor{\bsnm{Gramacy}, \binits{R.B.}}:
\bbtitle{Surrogates: {G}aussian Process Modeling, Design and Optimization for the Applied Sciences}.
\bpublisher{Chapman Hall/CRC},
\blocation{Boca Raton, Florida}
(\byear{2020}).
\bcomment{\url{http://bobby.gramacy.com/surrogates/}}
\end{bbook}
\endbibitem

\bibitem[\protect\citeauthoryear{Williams and Rasmussen}{2006}]{williams2006gaussian}
\begin{bbook}
\bauthor{\bsnm{Williams}, \binits{C.K.}},
\bauthor{\bsnm{Rasmussen}, \binits{C.E.}}:
\bbtitle{Gaussian Processes for Machine Learning}
vol. \bseriesno{2}.
\bpublisher{MIT press},
\blocation{Cambridge, MA}
(\byear{2006})
\end{bbook}
\endbibitem

\bibitem[\protect\citeauthoryear{Merrill et~al.}{2021}]{merrill2021empirical}
\begin{barticle}
\bauthor{\bsnm{Merrill}, \binits{E.}},
\bauthor{\bsnm{Fern}, \binits{A.}},
\bauthor{\bsnm{Fern}, \binits{X.}},
\bauthor{\bsnm{Dolatnia}, \binits{N.}}:
\batitle{An empirical study of {B}ayesian optimization: acquisition versus partition}.
\bjtitle{The Journal of Machine Learning Research}
\bvolume{22}(\bissue{1}),
\bfpage{200}--\blpage{224}
(\byear{2021})
\end{barticle}
\endbibitem

\bibitem[\protect\citeauthoryear{Gramacy and Apley}{2015}]{gramacy2015local}
\begin{barticle}
\bauthor{\bsnm{Gramacy}, \binits{R.B.}},
\bauthor{\bsnm{Apley}, \binits{D.W.}}:
\batitle{Local gaussian process approximation for large computer experiments}.
\bjtitle{Journal of Computational and Graphical Statistics}
\bvolume{24}(\bissue{2}),
\bfpage{561}--\blpage{578}
(\byear{2015})
\end{barticle}
\endbibitem

\bibitem[\protect\citeauthoryear{Vecchia}{1988}]{vecchia1988estimation}
\begin{barticle}
\bauthor{\bsnm{Vecchia}, \binits{A.V.}}:
\batitle{Estimation and model identification for continuous spatial processes}.
\bjtitle{Journal of the Royal Statistical Society Series B: Statistical Methodology}
\bvolume{50}(\bissue{2}),
\bfpage{297}--\blpage{312}
(\byear{1988})
\end{barticle}
\endbibitem

\bibitem[\protect\citeauthoryear{Katzfuss et~al.}{2020}]{katzfuss2020vecchia}
\begin{barticle}
\bauthor{\bsnm{Katzfuss}, \binits{M.}},
\bauthor{\bsnm{Guinness}, \binits{J.}},
\bauthor{\bsnm{Gong}, \binits{W.}},
\bauthor{\bsnm{Zilber}, \binits{D.}}:
\batitle{Vecchia approximations of gaussian-process predictions}.
\bjtitle{Journal of Agricultural, Biological and Environmental Statistics}
\bvolume{25},
\bfpage{383}--\blpage{414}
(\byear{2020})
\end{barticle}
\endbibitem

\bibitem[\protect\citeauthoryear{Furrer et~al.}{2006}]{furrer2006covariance}
\begin{barticle}
\bauthor{\bsnm{Furrer}, \binits{R.}},
\bauthor{\bsnm{Genton}, \binits{M.G.}},
\bauthor{\bsnm{Nychka}, \binits{D.}}:
\batitle{Covariance tapering for interpolation of large spatial datasets}.
\bjtitle{Journal of Computational and Graphical Statistics}
\bvolume{15}(\bissue{3}),
\bfpage{502}--\blpage{523}
(\byear{2006})
\end{barticle}
\endbibitem

\bibitem[\protect\citeauthoryear{Rahimi and Recht}{2007}]{rahimi2007random}
\begin{botherref}
\oauthor{\bsnm{Rahimi}, \binits{A.}},
\oauthor{\bsnm{Recht}, \binits{B.}}:
Random features for large-scale kernel machines.
Advances in neural information processing systems
\textbf{20}
(2007)
\end{botherref}
\endbibitem

\bibitem[\protect\citeauthoryear{Williams and Seeger}{2000}]{williams2000using}
\begin{botherref}
\oauthor{\bsnm{Williams}, \binits{C.}},
\oauthor{\bsnm{Seeger}, \binits{M.}}:
Using the nystr{\"o}m method to speed up kernel machines.
Advances in neural information processing systems
\textbf{13}
(2000)
\end{botherref}
\endbibitem

\bibitem[\protect\citeauthoryear{Snelson and Ghahramani}{2005}]{snelson2005sparse}
\begin{botherref}
\oauthor{\bsnm{Snelson}, \binits{E.}},
\oauthor{\bsnm{Ghahramani}, \binits{Z.}}:
Sparse gaussian processes using pseudo-inputs.
Advances in neural information processing systems
\textbf{18}
(2005)
\end{botherref}
\endbibitem

\bibitem[\protect\citeauthoryear{Titsias}{2009}]{titsias2009variational}
\begin{bchapter}
\bauthor{\bsnm{Titsias}, \binits{M.}}:
\bctitle{Variational learning of inducing variables in sparse gaussian processes}.
In: \bbtitle{Artificial Intelligence and Statistics},
pp. \bfpage{567}--\blpage{574}
(\byear{2009}).
\bcomment{PMLR}
\end{bchapter}
\endbibitem

\bibitem[\protect\citeauthoryear{Hensman et~al.}{2013}]{hensman2013gaussian}
\begin{bchapter}
\bauthor{\bsnm{Hensman}, \binits{J.}},
\bauthor{\bsnm{Fusi}, \binits{N.}},
\bauthor{\bsnm{Lawrence}, \binits{N.D.}}:
\bctitle{Gaussian processes for big data}.
In: \bbtitle{Proceedings of the Twenty-Ninth Conference on Uncertainty in Artificial Intelligence}.
\bsertitle{UAI'13},
pp. \bfpage{282}--\blpage{290}.
\bpublisher{AUAI Press},
\blocation{Arlington, Virginia, USA}
(\byear{2013})
\end{bchapter}
\endbibitem

\bibitem[\protect\citeauthoryear{Kandasamy et~al.}{2018}]{kandasamy2018parallelised}
\begin{bchapter}
\bauthor{\bsnm{Kandasamy}, \binits{K.}},
\bauthor{\bsnm{Krishnamurthy}, \binits{A.}},
\bauthor{\bsnm{Schneider}, \binits{J.}},
\bauthor{\bsnm{P{\'o}czos}, \binits{B.}}:
\bctitle{Parallelised {Bayesian} optimisation via thompson sampling}.
In: \bbtitle{International Conference on Artificial Intelligence and Statistics},
pp. \bfpage{133}--\blpage{142}
(\byear{2018}).
\bcomment{PMLR}
\end{bchapter}
\endbibitem

\bibitem[\protect\citeauthoryear{Eriksson et~al.}{2019}]{eriksson2019scalable}
\begin{bchapter}
\bauthor{\bsnm{Eriksson}, \binits{D.}},
\bauthor{\bsnm{Pearce}, \binits{M.}},
\bauthor{\bsnm{Gardner}, \binits{J.}},
\bauthor{\bsnm{Turner}, \binits{R.D.}},
\bauthor{\bsnm{Poloczek}, \binits{M.}}:
\bctitle{Scalable global optimization via local {Bayesian} optimization}.
In: \bbtitle{Advances in Neural Information Processing Systems},
pp. \bfpage{5496}--\blpage{5507}
(\byear{2019})
\end{bchapter}
\endbibitem

\bibitem[\protect\citeauthoryear{Gramacy et~al.}{2022}]{gramacy2022triangulation}
\begin{barticle}
\bauthor{\bsnm{Gramacy}, \binits{R.B.}},
\bauthor{\bsnm{Sauer}, \binits{A.}},
\bauthor{\bsnm{Wycoff}, \binits{N.}}:
\batitle{Triangulation candidates for {Bayesian} optimization}.
\bjtitle{Advances in Neural Information Processing Systems}
\bvolume{35},
\bfpage{35933}--\blpage{35945}
(\byear{2022})
\end{barticle}
\endbibitem

\bibitem[\protect\citeauthoryear{Jones et~al.}{1998}]{jones1998efficient}
\begin{barticle}
\bauthor{\bsnm{Jones}, \binits{D.R.}},
\bauthor{\bsnm{Schonlau}, \binits{M.}},
\bauthor{\bsnm{Welch}, \binits{W.J.}}:
\batitle{Efficient global optimization of expensive black-box functions}.
\bjtitle{Journal of Global optimization}
\bvolume{13},
\bfpage{455}--\blpage{492}
(\byear{1998})
\end{barticle}
\endbibitem

\bibitem[\protect\citeauthoryear{Garnett}{2023}]{garnett2023bayesian}
\begin{bbook}
\bauthor{\bsnm{Garnett}, \binits{R.}}:
\bbtitle{{Bayesian} Optimization}.
\bpublisher{Cambridge University Press}, \blocation{???}
(\byear{2023})
\end{bbook}
\endbibitem

\bibitem[\protect\citeauthoryear{Shahriari et~al.}{2015}]{shahriari2015taking}
\begin{barticle}
\bauthor{\bsnm{Shahriari}, \binits{B.}},
\bauthor{\bsnm{Swersky}, \binits{K.}},
\bauthor{\bsnm{Wang}, \binits{Z.}},
\bauthor{\bsnm{Adams}, \binits{R.P.}},
\bauthor{\bsnm{De~Freitas}, \binits{N.}}:
\batitle{Taking the human out of the loop: A review of {Bayesian} optimization}.
\bjtitle{Proceedings of the IEEE}
\bvolume{104}(\bissue{1}),
\bfpage{148}--\blpage{175}
(\byear{2015})
\end{barticle}
\endbibitem

\bibitem[\protect\citeauthoryear{Zhan and Xing}{2020}]{zhan2020expected}
\begin{barticle}
\bauthor{\bsnm{Zhan}, \binits{D.}},
\bauthor{\bsnm{Xing}, \binits{H.}}:
\batitle{Expected improvement for expensive optimization: a review}.
\bjtitle{Journal of Global Optimization}
\bvolume{78}(\bissue{3}),
\bfpage{507}--\blpage{544}
(\byear{2020})
\end{barticle}
\endbibitem

\bibitem[\protect\citeauthoryear{Ament et~al.}{2023}]{ament2023unexpected}
\begin{botherref}
\oauthor{\bsnm{Ament}, \binits{S.}},
\oauthor{\bsnm{Daulton}, \binits{S.}},
\oauthor{\bsnm{Eriksson}, \binits{D.}},
\oauthor{\bsnm{Balandat}, \binits{M.}},
\oauthor{\bsnm{Bakshy}, \binits{E.}}:
Unexpected improvements to expected improvement for bayesian optimization.
arXiv preprint arXiv:2310.20708
(2023)
\end{botherref}
\endbibitem

\bibitem[\protect\citeauthoryear{Franey et~al.}{2011}]{franey2011branch}
\begin{barticle}
\bauthor{\bsnm{Franey}, \binits{M.}},
\bauthor{\bsnm{Ranjan}, \binits{P.}},
\bauthor{\bsnm{Chipman}, \binits{H.}}:
\batitle{Branch and bound algorithms for maximizing expected improvement functions}.
\bjtitle{Journal of statistical planning and inference}
\bvolume{141}(\bissue{1}),
\bfpage{42}--\blpage{55}
(\byear{2011})
\end{barticle}
\endbibitem

\bibitem[\protect\citeauthoryear{Wang et~al.}{2020}]{wang2020learning}
\begin{barticle}
\bauthor{\bsnm{Wang}, \binits{L.}},
\bauthor{\bsnm{Fonseca}, \binits{R.}},
\bauthor{\bsnm{Tian}, \binits{Y.}}:
\batitle{Learning search space partition for black-box optimization using monte carlo tree search}.
\bjtitle{Advances in Neural Information Processing Systems}
\bvolume{33},
\bfpage{19511}--\blpage{19522}
(\byear{2020})
\end{barticle}
\endbibitem

\bibitem[\protect\citeauthoryear{Eriksson and Poloczek}{2021}]{eriksson2021scalable}
\begin{bchapter}
\bauthor{\bsnm{Eriksson}, \binits{D.}},
\bauthor{\bsnm{Poloczek}, \binits{M.}}:
\bctitle{Scalable constrained {Bayesian} optimization}.
In: \bbtitle{International Conference on Artificial Intelligence and Statistics},
pp. \bfpage{730}--\blpage{738}
(\byear{2021}).
\bcomment{PMLR}
\end{bchapter}
\endbibitem

\bibitem[\protect\citeauthoryear{Daulton et~al.}{2022}]{daulton2022multi}
\begin{bchapter}
\bauthor{\bsnm{Daulton}, \binits{S.}},
\bauthor{\bsnm{Eriksson}, \binits{D.}},
\bauthor{\bsnm{Balandat}, \binits{M.}},
\bauthor{\bsnm{Bakshy}, \binits{E.}}:
\bctitle{Multi-objective bayesian optimization over high-dimensional search spaces}.
In: \bbtitle{Uncertainty in Artificial Intelligence},
pp. \bfpage{507}--\blpage{517}
(\byear{2022}).
\bcomment{PMLR}
\end{bchapter}
\endbibitem

\bibitem[\protect\citeauthoryear{Sobol'}{1967}]{sobol1967distribution}
\begin{barticle}
\bauthor{\bsnm{Sobol'}, \binits{I.M.}}:
\batitle{On the distribution of points in a cube and the approximate evaluation of integrals}.
\bjtitle{Zhurnal Vychislitel'noi Matematiki i Matematicheskoi Fiziki}
\bvolume{7}(\bissue{4}),
\bfpage{784}--\blpage{802}
(\byear{1967})
\end{barticle}
\endbibitem

\bibitem[\protect\citeauthoryear{Mckay et~al.}{1979}]{Mckay:1979}
\begin{barticle}
\bauthor{\bsnm{Mckay}, \binits{D.}},
\bauthor{\bsnm{Beckman}, \binits{R.}},
\bauthor{\bsnm{Conover}, \binits{W.}}:
\batitle{A comparison of three methods for selecting vales of input variables in the analysis of output from a computer code}.
\bjtitle{Technometrics}
\bvolume{21},
\bfpage{239}--\blpage{245}
(\byear{1979})
\end{barticle}
\endbibitem

\bibitem[\protect\citeauthoryear{Lin and Tang}{2015}]{lin2015latin}
\begin{botherref}
\oauthor{\bsnm{Lin}, \binits{C.}},
\oauthor{\bsnm{Tang}, \binits{B.}}:
Latin hypercubes and space-filling designs.
Handbook of Design and Analysis of Experiments,
593--625
(2015)
\end{botherref}
\endbibitem

\bibitem[\protect\citeauthoryear{Bates and Pronzato}{2001}]{bates2001tri}
\begin{bchapter}
\bauthor{\bsnm{Bates}, \binits{R.}},
\bauthor{\bsnm{Pronzato}, \binits{L.}}:
\bctitle{Emulator-based global optimisation using lattices and delaunay tesselation}.
In: \bbtitle{Sensitivity Analysis of Model Output},
\bconflocation{Madrid (Espagne)},
pp. \bfpage{189}--\blpage{192}
(\byear{2001})
\end{bchapter}
\endbibitem

\bibitem[\protect\citeauthoryear{Barber et~al.}{1996}]{quickhull}
\begin{barticle}
\bauthor{\bsnm{Barber}, \binits{C.B.}},
\bauthor{\bsnm{Dobkin}, \binits{D.P.}},
\bauthor{\bsnm{Huhdanpaa}, \binits{H.}}:
\batitle{The quickhull algorithm for convex hulls}.
\bjtitle{ACM Trans. Math. Softw.}
\bvolume{22}(\bissue{4}),
\bfpage{469}--\blpage{483}
(\byear{1996})
\doiurl{10.1145/235815.235821}
\end{barticle}
\endbibitem

\bibitem[\protect\citeauthoryear{Schneider and Weil}{2008}]{schneider2008stochastic}
\begin{bbook}
\bauthor{\bsnm{Schneider}, \binits{R.}},
\bauthor{\bsnm{Weil}, \binits{W.}}:
\bbtitle{Stochastic and Integral Geometry}
vol. \bseriesno{1}.
\bpublisher{Springer}, \blocation{???}
(\byear{2008})
\end{bbook}
\endbibitem

\bibitem[\protect\citeauthoryear{Polianskii}{2022}]{polianskii2022breaking}
\begin{botherref}
\oauthor{\bsnm{Polianskii}, \binits{V.}}:
Breaking the dimensionality curse of voronoi tessellations.
PhD thesis,
KTH Royal Institute of Technology
(2022)
\end{botherref}
\endbibitem

\bibitem[\protect\citeauthoryear{Arya et~al.}{1998}]{arya1998optimal}
\begin{barticle}
\bauthor{\bsnm{Arya}, \binits{S.}},
\bauthor{\bsnm{Mount}, \binits{D.M.}},
\bauthor{\bsnm{Netanyahu}, \binits{N.S.}},
\bauthor{\bsnm{Silverman}, \binits{R.}},
\bauthor{\bsnm{Wu}, \binits{A.Y.}}:
\batitle{An optimal algorithm for approximate nearest neighbor searching fixed dimensions}.
\bjtitle{Journal of the ACM (JACM)}
\bvolume{45}(\bissue{6}),
\bfpage{891}--\blpage{923}
(\byear{1998})
\end{barticle}
\endbibitem

\bibitem[\protect\citeauthoryear{Bentley}{1975}]{bentley1975multidimensional}
\begin{barticle}
\bauthor{\bsnm{Bentley}, \binits{J.L.}}:
\batitle{Multidimensional binary search trees used for associative searching}.
\bjtitle{Communications of the ACM}
\bvolume{18}(\bissue{9}),
\bfpage{509}--\blpage{517}
(\byear{1975})
\end{barticle}
\endbibitem

\bibitem[\protect\citeauthoryear{Bull}{2011}]{bull2011convergence}
\begin{botherref}
\oauthor{\bsnm{Bull}, \binits{A.D.}}:
Convergence rates of efficient global optimization algorithms.
Journal of Machine Learning Research
\textbf{12}(10)
(2011)
\end{botherref}
\endbibitem

\bibitem[\protect\citeauthoryear{Kim and Sanz-Alonso}{2024}]{kim2024enhancing}
\begin{botherref}
\oauthor{\bsnm{Kim}, \binits{H.}},
\oauthor{\bsnm{Sanz-Alonso}, \binits{D.}}:
Enhancing {G}aussian process surrogates for optimization and posterior approximation via random exploration.
arXiv preprint arXiv:2401.17037
(2024)
\end{botherref}
\endbibitem

\bibitem[\protect\citeauthoryear{Gramacy}{2016}]{laGP}
\begin{barticle}
\bauthor{\bsnm{Gramacy}, \binits{R.B.}}:
\batitle{{laGP}: Large-scale spatial modeling via local approximate gaussian processes in {R}}.
\bjtitle{Journal of Statistical Software}
\bvolume{72}(\bissue{1}),
\bfpage{1}--\blpage{46}
(\byear{2016})
\doiurl{10.18637/jss.v072.i01}
\end{barticle}
\endbibitem

\bibitem[\protect\citeauthoryear{Kok and Sandrock}{2009}]{kok2009locating}
\begin{barticle}
\bauthor{\bsnm{Kok}, \binits{S.}},
\bauthor{\bsnm{Sandrock}, \binits{C.}}:
\batitle{Locating and characterizing the stationary points of the extended rosenbrock function}.
\bjtitle{Evolutionary computation}
\bvolume{17}(\bissue{3}),
\bfpage{437}--\blpage{453}
(\byear{2009})
\end{barticle}
\endbibitem

\bibitem[\protect\citeauthoryear{Brockman et~al.}{2016}]{openaigym}
\begin{botherref}
\oauthor{\bsnm{Brockman}, \binits{G.}},
\oauthor{\bsnm{Cheung}, \binits{V.}},
\oauthor{\bsnm{Pettersson}, \binits{L.}},
\oauthor{\bsnm{Schneider}, \binits{J.}},
\oauthor{\bsnm{Schulman}, \binits{J.}},
\oauthor{\bsnm{Tang}, \binits{J.}},
\oauthor{\bsnm{Zaremba}, \binits{W.}}:
OpenAI Gym
(2016)
\end{botherref}
\endbibitem

\bibitem[\protect\citeauthoryear{Wang et~al.}{2017}]{wang2017batched}
\begin{bchapter}
\bauthor{\bsnm{Wang}, \binits{Z.}},
\bauthor{\bsnm{Li}, \binits{C.}},
\bauthor{\bsnm{Jegelka}, \binits{S.}},
\bauthor{\bsnm{Kohli}, \binits{P.}}:
\bctitle{Batched high-dimensional {Bayesian} optimization via structural kernel learning}.
In: \bbtitle{International Conference on Machine Learning},
pp. \bfpage{3656}--\blpage{3664}
(\byear{2017}).
\bcomment{PMLR}
\end{bchapter}
\endbibitem

\bibitem[\protect\citeauthoryear{Wang et~al.}{2018}]{wang2018batched}
\begin{bchapter}
\bauthor{\bsnm{Wang}, \binits{Z.}},
\bauthor{\bsnm{Gehring}, \binits{C.}},
\bauthor{\bsnm{Kohli}, \binits{P.}},
\bauthor{\bsnm{Jegelka}, \binits{S.}}:
\bctitle{Batched large-scale {Bayesian} optimization in high-dimensional spaces}.
In: \bbtitle{International Conference on Artificial Intelligence and Statistics},
pp. \bfpage{745}--\blpage{754}
(\byear{2018}).
\bcomment{PMLR}
\end{bchapter}
\endbibitem

\bibitem[\protect\citeauthoryear{Smith et~al.}{2023}]{smith2023}
\begin{barticle}
\bauthor{\bsnm{Smith}, \binits{J.W.}},
\bauthor{\bsnm{Thomas}, \binits{R.Q.}},
\bauthor{\bsnm{Johnson}, \binits{L.R.}}:
\batitle{Parameterizing lognormal state space models using moment matching}.
\bjtitle{Environmental and Ecological Statistics}
(\byear{2023})
\doiurl{10.1007/s10651-023-00570-x}
\end{barticle}
\endbibitem

\bibitem[\protect\citeauthoryear{Ionides et~al.}{2015}]{IF2}
\begin{barticle}
\bauthor{\bsnm{Ionides}, \binits{E.L.}},
\bauthor{\bsnm{Nguyen}, \binits{D.}},
\bauthor{\bsnm{Atchadé}, \binits{Y.}},
\bauthor{\bsnm{Stoev}, \binits{S.}},
\bauthor{\bsnm{King}, \binits{A.A.}}:
\batitle{Inference for dynamic and latent variable models via iterated, perturbed bayes maps}.
\bjtitle{Proceedings of the National Academy of Sciences}
\bvolume{112}(\bissue{3}),
\bfpage{719}--\blpage{724}
(\byear{2015})
\doiurl{10.1073/pnas.1410597112}
{\href{https://arxiv.org/abs/https://www.pnas.org/doi/pdf/10.1073/pnas.1410597112}{{https://www.pnas.org/doi/pdf/10.1073/pnas.1410597112}}}
\end{barticle}
\endbibitem

\bibitem[\protect\citeauthoryear{King et~al.}{2008}]{king_ionides_dhaka}
\begin{barticle}
\bauthor{\bsnm{King}, \binits{A.}},
\bauthor{\bsnm{Ionides}, \binits{E.}},
\bauthor{\bsnm{Pascual}, \binits{M.}},
\bauthor{\bsnm{Bouma}, \binits{M.}}:
\batitle{Inapparent infections and cholera dynamics}.
\bjtitle{Nature}
\bvolume{454},
\bfpage{877}--\blpage{80}
(\byear{2008})
\doiurl{10.1038/nature07084}
\end{barticle}
\endbibitem

\bibitem[\protect\citeauthoryear{Ionides et~al.}{2011}]{ionides2011_if}
\begin{barticle}
\bauthor{\bsnm{Ionides}, \binits{E.L.}},
\bauthor{\bsnm{Bhadra}, \binits{A.}},
\bauthor{\bsnm{Atchad{\'e}}, \binits{Y.}},
\bauthor{\bsnm{King}, \binits{A.}}:
\batitle{{Iterated filtering}}.
\bjtitle{The Annals of Statistics}
\bvolume{39}(\bissue{3}),
\bfpage{1776}--\blpage{1802}
(\byear{2011})
\doiurl{10.1214/11-AOS886}
\end{barticle}
\endbibitem

\bibitem[\protect\citeauthoryear{Ionides et~al.}{2006}]{IF1}
\begin{barticle}
\bauthor{\bsnm{Ionides}, \binits{E.L.}},
\bauthor{\bsnm{Bretó}, \binits{C.}},
\bauthor{\bsnm{King}, \binits{A.A.}}:
\batitle{Inference for nonlinear dynamical systems}.
\bjtitle{Proceedings of the National Academy of Sciences}
\bvolume{103}(\bissue{49}),
\bfpage{18438}--\blpage{18443}
(\byear{2006})
\doiurl{10.1073/pnas.0603181103}
{\href{https://arxiv.org/abs/https://www.pnas.org/doi/pdf/10.1073/pnas.0603181103}{{https://www.pnas.org/doi/pdf/10.1073/pnas.0603181103}}}
\end{barticle}
\endbibitem

\bibitem[\protect\citeauthoryear{Andrieu et~al.}{2010}]{pMCMCAndrieu}
\begin{barticle}
\bauthor{\bsnm{Andrieu}, \binits{C.}},
\bauthor{\bsnm{Doucet}, \binits{A.}},
\bauthor{\bsnm{Holenstein}, \binits{R.}}:
\batitle{Particle markov chain monte carlo methods}.
\bjtitle{Journal of the Royal Statistical Society: Series B (Statistical Methodology)}
\bvolume{72}(\bissue{3}),
\bfpage{269}--\blpage{342}
(\byear{2010})
\doiurl{10.1111/j.1467-9868.2009.00736.x}
\end{barticle}
\endbibitem

\bibitem[\protect\citeauthoryear{Bhadra}{2010}]{bhadra2010}
\begin{botherref}
\oauthor{\bsnm{Bhadra}, \binits{A.}}:
Discussion of ‘particle markov chain monte carlo methods’ by c. andrieu, a. doucet and r. holenstein.
Journal of the Royal Statistical Society B,
314--315
(2010)
\doiurl{10.1111/j.1467-9868.2009.00736.x}
\end{botherref}
\endbibitem

\bibitem[\protect\citeauthoryear{Sauer et~al.}{2023}]{sauer2023active}
\begin{barticle}
\bauthor{\bsnm{Sauer}, \binits{A.}},
\bauthor{\bsnm{Gramacy}, \binits{R.B.}},
\bauthor{\bsnm{Higdon}, \binits{D.}}:
\batitle{Active learning for deep gaussian process surrogates}.
\bjtitle{Technometrics}
\bvolume{65}(\bissue{1}),
\bfpage{4}--\blpage{18}
(\byear{2023})
\end{barticle}
\endbibitem

\end{thebibliography}

\newpage
\appendix

\end{document}